\documentclass[journal]{IEEEtran}

\usepackage{comment}
\renewcommand{\baselinestretch}{0.92}
\usepackage{changepage}
\usepackage{amsmath}
\usepackage{amssymb}
\usepackage[dvips]{graphicx}
\usepackage{setspace}
\usepackage{epsfig}
\usepackage{color}
\usepackage{morefloats}
\usepackage{enumerate}
\usepackage{bbm}
\usepackage{cases}
\usepackage{enumitem}
\usepackage{booktabs}
\usepackage{graphicx}
\usepackage[caption=false]{subfig}
\allowdisplaybreaks

\usepackage{amsfonts} % For \mathbb
\usepackage{bm} % For bold math symbols
\usepackage{comment}
\usepackage{caption}
\usepackage{subcaption}
\newcommand{\Ztb}{\tilde{\boldsymbol{Z}}}
\newcommand{\Atb}{\tilde{\boldsymbol{A}}}

\newcommand{\Ytb}{\hat{\boldsymbol{y}}}
\usepackage{subfig}
\usepackage{float}        % for precise figure placement, if needed
\usepackage{stfloats} % Allows figure* at the top of page in IEEEt

\usepackage[a4paper,
            bindingoffset=0.2in,
            left=0.6in,
            right=0.6in,
            top=0.7in,
            bottom=0.9in,
            footskip=.25in]{geometry}

\begin{document}
\renewcommand{\baselinestretch}{0.85}
% \pagestyle{empty}
% Before the part you want to change

% paper title
% can use linebreaks \\ within to get better formatting as desired
% Do not put math or special symbols in the title.
% \title{Computationally Efficient Transformers for Precise Device Localization in NLoS Environment with Distributed Sensors}
%\title{SST: Sensor Snapshot Tokenization for Transformer-Based Indoor Wireless Localization in NLOS Environments}
\title{Transforming Indoor Localization: Advanced Transformer Architecture for NLOS Dominated Wireless Environments with Distributed Sensors}
%Optimal Channel Switching and Randomization over Flat-Fading Channels for \textcolor{blue}{Outage} Capacity Maximization
\author{Saad Masrur\vspace{-0.5cm}}
\author{Saad Masrur$^1$, Jung-Fu (Thomas) Cheng$^2$, Atieh R. Khamesi$^2$,  \.{I}smail G\"{u}ven\c{c}$^1$\\
$^1$Department of Electrical and Computer Engineering, North Carolina State University, Raleigh, NC, USA\\
$^2$Ericsson Research, Santa Clara, CA, USA\\
{\texttt \{smasrur,iguvenc\}@ncsu.edu, \texttt  \{atieh.rajabi.khamesi,thomas.cheng\}@ericsson.com}

}

% \author{{Saad Masrur\thanks{S. Masrur and S. Gezici are with the Dept. of Electrical and Electronics Eng., Bilkent Univ., Bilkent, Ankara, Turkey, Tel: +90 (312) 290-3139 (e-mails: saad@ee.bilkent.edu.tr, gezici@ee.bilkent.edu.tr).} and Sinan Gezici, {\emph{Senior Member, IEEE}}
% \vspace{-0.6cm}
% }}

% The paper headers
%\markboth{Journal of \LaTeX\ Class Files,~Vol.~11, No.~4, December~2012}%
%{Shell \MakeLowercase{\textit{et al.}}: Bare Demo of IEEEtran.cls for Journals}
% The only time the second header will appear is for the odd-numbered pages
% after the title page when using the twoside option. 

% make the title area

\maketitle
\thispagestyle{empty}
% As a general rule, do not put math, special symbols or citations
% in the abstract or keywords.
 % and the number of parameters
\begin{abstract}
\pagestyle{empty}
%Indoor localization in highly Non-line-of-sight (NLOS) environments poses considerable challenges, often resulting in significant inaccuracies for conventional algorithms. Although deep learning has been applied to tackle this issue, many approaches fail to account for computational complexity, particularly in terms of floating point operations (FLOPs), rendering them unsuitable for resource-limited devices. Conventional tokenization often amalgamates multiple variables, such as delayed events and various physical measurements, which can impede the model’s ability to learn variable-specific representations and can result in less meaningful attention maps. In this paper, we analyze the behavior of Transformers in almost NLOS environments for localization and introduce an efficient tokenization technique aimed at addressing both computational and data constraints. 
%Conventional tokenization often amalgamates multiple variables, such as delayed events and various physical measurements, which can dilute variable-specific representations and result in less informative attention maps.
%  
%This highlights the need for advanced models that can effectively operate under NLOS conditions while maintaining a low computational budget.
%and the number of parameters
Indoor localization in challenging non-line-of-sight (NLOS) environments often leads to poor accuracy with traditional approaches. Deep learning (DL) has been applied to tackle these challenges; however, many DL approaches overlook computational complexity, especially for floating-point operations (FLOPs), making them unsuitable for resource-limited devices.  
Transformer-based models have achieved remarkable success in natural language processing (NLP) and computer vision (CV) tasks, motivating their use in wireless applications. However, their use in indoor localization remains nascent, and directly applying Transformers for indoor localization can be both computationally intensive and exhibit limitations in accuracy.
To address these challenges, in this work, we introduce a novel tokenization approach, referred to as Sensor Snapshot Tokenization (\emph{SST}), which preserves variable-specific representations of power delay profile (PDP) and enhances attention mechanisms by effectively capturing multi-variate correlation. Complementing this, we propose a lightweight Swish-Gated Linear Unit-based Transformer (\emph{L-SwiGLU-T}) model, designed to reduce computational complexity without compromising localization accuracy. Together, these contributions mitigate the computational burden and dependency on large datasets, making Transformer models more efficient and suitable for resource-constrained scenarios. Experimental results on simulated and real-world datasets demonstrate that \emph{SST} and \emph{L-SwiGLU-T} achieve substantial accuracy and efficiency gains, outperforming larger Transformer and CNN baselines by over 40\% while using significantly fewer FLOPs and training samples.

\pagestyle{empty}

\textit{Index~Terms}--- 3GPP InF, FLOPs, indoor localization, NLOS, signal processing, tokenization, Transformer. 
\end{abstract}
%\comment{In the context of wireless communication, the challenge is further compounded by the typically smaller datasets available, compared to those in computer vision (CV), and natural language processing (NLP).}

%\begin{IEEEkeywords}
%Channel switching, jamming, Nash equilibrium, capacity, time-sharing, power allocation.
%\end{IEEEkeywords}

% For peerreview papers, this IEEEtran command inserts a page break and
% creates the second title. It will be ignored for other modes.
%\IEEEpeerreviewmaketitle

\vspace{-0.3cm}

\section{Introduction}\label{sec:intro}
\pagestyle{empty}
% The evolution of wireless communication technologies is progressing rapidly, with 5G on the verge of deployment and discussions around 6G gaining momentum. This exponential growth in connected devices will introduce groundbreaking location-based services such as vehicle-to-everything (V2X) communication, urban air mobility (UAM), multisensory extended reality (XR), near-real-time robotic operations, and Smart X applications \cite{lee2023towards,nguyen2019localization}. These innovations will necessitate the extensive deployment of Internet of Things (IoT) infrastructures. Consequently, there is a growing demand for robust and precise localization techniques with reduced computational complexity to meet the quality of service (QoS) for these services. The need for reduced complexity arises because most indoor positioning devices lack sufficient computing power. Effective indoor localization is crucial for a wide range of applications, including navigation, safety and rescue operations, and efficient resource allocation for customized user services.

The rapid advancement of wireless communication technologies, fueled by the deployment of 5G and the emerging vision of 6G, is driving transformative innovations in location-based services. Applications such as vehicle-to-everything (V2X) communication, urban air mobility (UAM), extended reality (XR), near-real-time robotic operations, and Smart X systems \cite{lee2023towards,nguyen2019localization} are at the forefront of this technological evolution. These advancements require the large-scale integration of Internet of Things (IoT) infrastructures, thereby intensifying the demand for accurate and efficient localization methods with reduced computational complexity to ensure the stringent quality of service (QoS) necessary for these cutting-edge applications. Indoor localization is particularly critical for applications such as navigation, safety and rescue operations, and resource optimization for customized user services. However, most indoor positioning devices remain constrained by limited computing power, highlighting the need for localization methods that are both efficient and accurate while supporting the demands of next-generation wireless systems.

% However, the need for reduced computational complexity is paramount, as most indoor positioning devices are constrained by limited computing power. Addressing these challenges requires the development of efficient and precise localization methods capable of operating within these constraints while supporting the demands of modern wireless communication systems.

Global Navigation Satellite Systems (GNSS), commonly used for outdoor localization, face significant challenges in indoor environments due to severe channel conditions such as shadowing, fading, and noise, as well as the high probability of non-line-of-sight (NLOS) situations. Conventional wireless localization approaches often rely on signal-level metrics such as Time of Arrival (ToA), Time Difference of Arrival (TDoA), Time of Flight (ToF), Angle of Arrival (AoA), and Received Signal Strength Indicator (RSSI) \cite{singh2021machine}. These metrics are commonly used in geometric or statistical localization methods, which perform poorly in NLOS-rich indoor environments such as factories, hospitals, and shopping malls due to their reliance on complex, environment-dependent empirical models.

Fingerprinting \cite{shang2022overview} is a widely used indoor localization approach that leverages signal measurements such as Received Signal Strength Indicator (RSSI) or Channel State Information (CSI). However, traditional machine-learning methods struggle with environmental variability and fail to capture complex spatial–temporal dependencies. Deep learning (DL) models—such as Convolutional Neural Networks (CNNs) and Multi-Layer Perceptrons (MLPs) \cite{singh2021machine} learn more discriminative features and offer improved robustness, but they also increase system complexity and computational cost, hindering real-time implementation and scalability. Therefore, this work aims to achieve a balance between accuracy and efficiency, enabling adaptable and lightweight fingerprinting-based localization.

The Transformer model \cite{vaswani2017attention} offers several advantages over traditional sequence processing techniques such as Recurrent Neural Networks (RNNs) and Long Short-Term Memory (LSTM) networks, including parallelization capability, reduced training time, and improved handling of long-range dependencies \cite{khan2022transformers}.
Transformers, despite their proven success in fields such as natural language processing (NLP), computer vision (CV), and machine translation \cite{vaswani2017attention, dosovitskiy2020image}, have not been extensively used for indoor localization. This underutilization is largely due to the lack of inductive biases such as locality and translation invariance, which are naturally present in CNNs but absent in Transformers, and due to the Transformer's dependence on large datasets to capture both local and global features \cite{dosovitskiy2020image}. These challenges are further amplified by the way inputs are tokenized, which can impact the model’s training effectiveness.

% In the wireless communication field, we lack access to large datasets, and collecting data of this magnitude is both computationally expensive and cost-inefficient. \textcolor{red}{A multi-antenna system in highly complex indoor wireless environments faces significant challenges due to the presence of multipath propagation, where signals reflect, diffract, and scatter off walls, furniture, and other obstacles. This results in signal superposition, leading to phase shifts, fading, and interference.when trying to resolve spatial and temporal variations in such environments.}  Furthermore, in the context of indoor localization, the existing network infrastructure does not possess the high computational power required to test models with billions of  FLOPs and parameters.
% \textcolor{red}{This highlights the need for a more efficient approach to reduce the reliance of Transformers on large datasets while simultaneously lowering their computational complexity.}\\
In highly NLOS indoor wireless environments, multi-antenna systems face significant challenges due to multipath propagation, where signals undergo reflection, diffraction, and scattering from walls, furniture, and other obstacles. This complex propagation leads to signal superposition, causing phase shifts, fading, and interference, which further complicate the resolution of spatial and temporal variations in such environments. Moreover, in the context of indoor localization, collecting large-scale localization datasets is costly and time-consuming, while existing infrastructures lack the computational capacity to test models requiring billions of FLOPs. These challenges underscore the need for more efficient approaches that reduce the dependence of Transformer-based models on large datasets while simultaneously lowering their computational complexity, enabling practical deployment in indoor wireless environments.

% Moreover, in the context of indoor localization, access to large-scale datasets is limited, as collecting data of this magnitude is both computationally expensive and cost-inefficient. Additionally, existing network infrastructures lack the high computational resources necessary to test models with billions of FLOPs, which further constrains the deployment of computationally intensive models.  These challenges underscore the need for more efficient approaches that reduce the dependence of Transformer-based models on large datasets while simultaneously lowering their computational complexity, enabling practical deployment in complex indoor wireless environments.

One critical yet often overlooked aspect of Transformers, particularly within the wireless communication domain, is tokenization. Conventional tokenization methods typically aggregate multiple variables, such as delayed events and physical measurements, into a single representation. This aggregation can obscure variable-specific features, leading to less informative attention maps and degrading model performance. Consequently, the model's dependency on extensive datasets increases. Therefore, in this paper, we investigate the Transformer architecture and enhance its learning capabilities by proposing a sophisticated and physically interpretable tokenization technique referred to as \emph{Sensor Snapshot Tokenization (SST)} that leverages the characteristics of wireless communication systems, specifically addressing inherent channel independence and enabling the Transformer to effectively learn the multivariate correlations using the tokens form multi-antenna radio system. SST produces variate-centric representations with more meaningful attention maps, reducing the need for large models and extensive training data. To further enhance efficiency and reduce computational complexity, we introduce a lightweight Transformer variant, termed the Lightweight Swish-Gated Linear Unit Transformer (\emph{L-SwiGLU-T}). Building upon the Vanilla Transformer, \emph{L-SwiGLU-T} incorporates selective normalization, compact feed-forward design, and an efficient prediction mechanism, collectively improving stability and reducing FLOPs.

% This approach facilitates the development of variate-centric representations, resulting in meaningful attention maps and leading to a decreased dependence of the Transformer on large models and extensive datasets.

% To further enhance efficiency and reduce computational complexity, we introduced modifications to several components of the Vanilla Transformer (Vanilla-T) architecture. These include replacing conventional normalization layers, redesigning the feed-forward network, and redefining the prediction mechanism using tokens from the final encoder block. The resulting enhanced Transformer model is termed the Lightweight Swish-Gated Linear Unit-based Transformer {(\emph{L-SwiGLU-T}), optimized for improved performance in resource-constrained scenarios. Furthermore, we perform a comprehensive evaluation of various architectural configurations within Transformers across datasets of varying sizes. This analysis underscores the efficacy of the proposed methodology, illuminating potential avenues for further research and development.

The primary contributions of this work are summarized as follows:
\begin{itemize}
    \item We define indoor localization in multi-antenna systems under highly NLOS conditions as a multivariate modeling challenge, addressed using Transformers to capture complex channel dynamics.
    \item We propose a novel 
    %Sensor Snapshot Tokenization (\emph{SST}) 
    tokenization technique, \emph{SST}, which enables the Transformer to capture multivariate correlations and generate meaningful attention maps, significantly reducing reliance on large datasets.
    \item We propose \emph{L-SwiGLU-T}, a modified Transformer architecture that enhances computational efficiency and positioning accuracy by replacing the MLP-based feed-forward network (FFN) with a Swish gated linear unit (GLU)-based FFN. Additionally, we replace standard components, such as the class token and normalization layers, with more efficient counterparts. Furthermore, by introducing a global average pooling (GAP) layer, we demonstrate that positional embeddings can be omitted without adversely affecting positioning accuracy.
    \item We added a comparison of the proposed model against two recent lightweight Transformer architectures, and validated its performance on a real-world dataset.
    \item We conduct an extensive evaluation of various Transformer architectures across datasets and models of varying sizes, demonstrating the proposed methodology's superior accuracy and computational efficiency.
\end{itemize}

Our preliminary study in \cite{11162366} introduced the \emph{SST} method and demonstrated its initial performance improvements using vanilla Transformer. In contrast, this paper extends that work by introducing a lightweight Transformer architecture (L-SwiGLU-T), evaluating across multiple datasets and model sizes, and validating performance on real-world measurements.

The paper is organized as follows. Section \ref{section:LitRev} provides a comprehensive review of the existing literature. The system model is described in detail in Section \ref{sec:system}. Section \ref{section:TokenPre} explains the preprocessing steps and the tokenization methods. The design of transformer architectures for processing distributed sensor signals is discussed in Section \ref{section:DesignTransformer}. Extensive evaluation studies and analyses are presented in Section \ref{sec:Nume}. Finally, the paper concludes in Section \ref{Conclusion}.

\section{Literature Review} \label{section:LitRev}
In the literature, indoor positioning is classified into two major categories: geometric-based methods and fingerprinting-based methods \cite{singh2021machine}. Geometric approaches, such as trilateration and triangulation, use parameters such as AoA, TDoA, ToF, and ToA for positioning. Although these algorithms are well-established and extensively studied, they perform poorly in indoor scenarios due to outlier distortion caused by NLOS and multipath challenges \cite{singh2021machine}. %from refraction and reflection 

Conversely, fingerprinting methods construct a database of signal features collected from known locations to predict the position of new measurements. Compared with geometric approaches, they are simpler, easily integrated into smart devices, and can achieve good accuracy using existing wireless infrastructure \cite{singh2021machine, alhomayani2020deep}. Early ML-based fingerprinting primarily relied on shallow algorithms such as K-nearest neighbors (KNN), Decision Trees (DTs), Support Vector Machines (SVMs), and Random Forests, which are costly to train and perform poorly on large or complex datasets \cite{nasir2024hytra}. With the advent of deep learning, models such as Multilayer Perceptrons (MLPs), RNNs, LSTMs, and CNNs have been explored, offering improved feature learning and generalization. A comprehensive overview of DL-based fingerprinting, including model comparisons and input modalities, is provided in \cite{singh2021machine, alhomayani2020deep}.

% Conversely, fingerprinting methods involve creating a database by collecting signals from various locations and extracting features from them. This constructed database is then used to predict the location of new signals. Compared to geometric approaches, fingerprinting methods are relatively simple, easily integrated into smart devices, and capable of achieving acceptable accuracy with support from existing wireless infrastructure, which explains their widespread exploration in the literature \cite{singh2021machine, alhomayani2020deep}.

% Most of the work involving ML-based fingerprinting has utilized shallow ML algorithms like K-nearest neighbors (KNN), Decision Trees (DTs), Support Vector Machines (SVM), and Random Forests. These algorithms are computationally expensive to train on large datasets and their performance tends to degrade. Furthermore, the
% performance of traditional ML methods will not scale with larger, more complex datasets \cite{nasir2024hytra}. With the advent of DL models, researchers have begun exploring algorithms like $\mathsf{MLP}$s, RNNs, LSTMs, and CNNs for fingerprinting. Comprehensive details about deep learning-based fingerprinting methods, including the pros and cons of different methods and various types of inputs used for positioning, are provided in \cite{singh2021machine, alhomayani2020deep}. 

Numerous DL techniques, including CNNs and LSTMs, have achieved remarkable success in indoor localization \cite{hoang2020cnn, gao2022toward}. In \cite{cheng2024}, a fully complex-valued neural network (CVNN) was proposed for positioning in environments with moderate Line-of-Sight (LoS) conditions, demonstrating high accuracy in 2D localization tasks. Building on this work, \cite{khamesi2024} extended the approach by employing a complex-valued ResNet model tailored for highly NLoS scenarios.

With the growing success of Transformer architectures, several studies have explored their use in indoor localization. In \cite{li2023lot}, a Vision Transformer (ViT) \cite{dosovitskiy2020image}, an encoder-only Transformer, was applied to fingerprinting-based localization, where a CNN generated CSI patches that were processed by Transformer encoders. Similarly, \cite{zhou2024vtil} employed ViT for WiFi-assisted localization, using Principal Component Analysis (PCA) for input normalization and converting RSSI data into grayscale images, achieving a 50th-percentile error of $1.788$~m. However, PCA-based preprocessing is impractical for deployment since the transformation cannot be consistently applied during inference, leading to degraded performance. Moreover, the dataset used in \cite{zhou2024vtil} is not publicly available, limiting reproducibility and evaluation under different LoS/NLoS conditions.

% With the recent success of Transformer-based models in various applications, researchers have become increasingly interested in exploring their potential for indoor localization. In \cite{li2023lot}, the Vision Transformer (ViT) \cite{dosovitskiy2020image}, which \textcolor{red}{is} encoder-only Transformer model, was employed for fingerprinting-based indoor localization. The approach involved using a CNN to create patches from the CSI matrix, which were then fed into the Transformer encoder blocks for localization. Similarly, the authors in \cite{zhou2024vtil} used ViT for WiFi-assisted indoor localization. They applied Principal Component Analysis (PCA) for input normalization and created RSSI gray images from the normalized RSSI data, achieving a $50^{th}$ percentile error of $1.788$~m. However, the use of PCA makes this approach impractical for real-world applications, as the same transformation cannot be consistently applied during testing, leading to performance degradation. Additionally, the dataset used in this study is not publicly available, making it difficult to assess the proportion of LoS.

In \cite{nasir2024hytra}, a Transformer encoder-only network is utilized for Received Signal Strength (RSS) based WiFi fingerprinting, where the values from wireless access points (WAPs) serve as inputs. The study focuses on building floor and room prediction tasks using RSS data. However, since only a single RSS value per WAP is available, the model's ability to extract meaningful spatial or temporal features is inherently limited. Furthermore, the use of PCA as a pre-processing step may not be optimal.

The authors in \cite{lutakamale2024hybrid} proposed a hybrid model incorporating CNNs and ViT for localization in Long Range Wide Area Network (LoRaWAN). In this model, CNN is used to learn local features while ViT captures global features. Similarly, in \cite{prasad2024vision}, the authors presented a CNN-aided ViT-based indoor localization method. In this approach, the CSI matrix is treated as an image and fed into a CNN, with the resulting features used as patches for the ViT. In the context of indoor localization for MIMO systems, the study in \cite{xu2024swin} adopts a similar hybrid approach, utilizing a variant of the ViT alongside a CNN for feature extraction.

An approach described in \cite{pan2022transformer} employs a Transformer for signal source localization using the AoA, creating images based on the reference point's location and arrival angle. However, the process for generating these images is not explained. Additionally, this method utilizes both the encoder and decoder parts of the Transformer, which is unusual since typically only the encoder is used for classification and regression tasks. The work and rationale for this architectural choice are also not provided. ViT have also been investigated for human activity recognition (HAR) using WiFi CSI \cite{luo2024vision}. In this approach, the raw CSI data is transformed into a CSI spectrogram, which is then used as input to the ViT model for effective HAR.

In the studies discussed, the input data was predominantly treated as images, with ViTs employed as localization algorithms without any modifications to their architecture. Although this approach capitalizes on ViT's inherent strengths in processing image data, it fails to fully exploit the unique characteristics of wireless communication systems. Specifically, wireless environments exhibit phenomena such as channel independence, signal fading, and multipath propagation, which are critical for accurate localization. By treating the data solely as visual inputs, these approaches overlook the rich temporal and spatial correlations that are fundamental to wireless communication. Ineffective utilization of the transformer architecture results in an increased reliance on larger Transformer models and necessitates the use of extensive datasets. Furthermore, in all these studies, the computational aspects of the resource-intensive Transformer models are largely overlooked, highlighting a significant challenge for their implementation in real-time, resource-constrained environments.

%In the wireless communication field, we lack access to such large datasets, and collecting data of this magnitude is both computationally expensive and cost-inefficient. Specifically in the context of indoor localization, the existing network infrastructure does not possess the high computational power required to test models with billions of parameters. Therefore, in this paper, we propose a sophisticated and physically interpretable tokenization technique that leverages the characteristics of wireless communication systems, specifically addressing inherent channel independence and enabling the transformer to effectively learn the multivariate correlations using  the tokens. This approach facilitates the development of variate-centric representations, resulting in meaningful attention maps and leading to a decreased dependence of the transformer on large models and extensive datasets.

%to achieve optimal performance.

%---------------------------------------------------------------------------------------------------------

\vspace{-0.1cm}

\section{System Model}\label{sec:system}
\begin{figure}
%\vspace{-0.5cm}
	\includegraphics[width=0.9\linewidth]{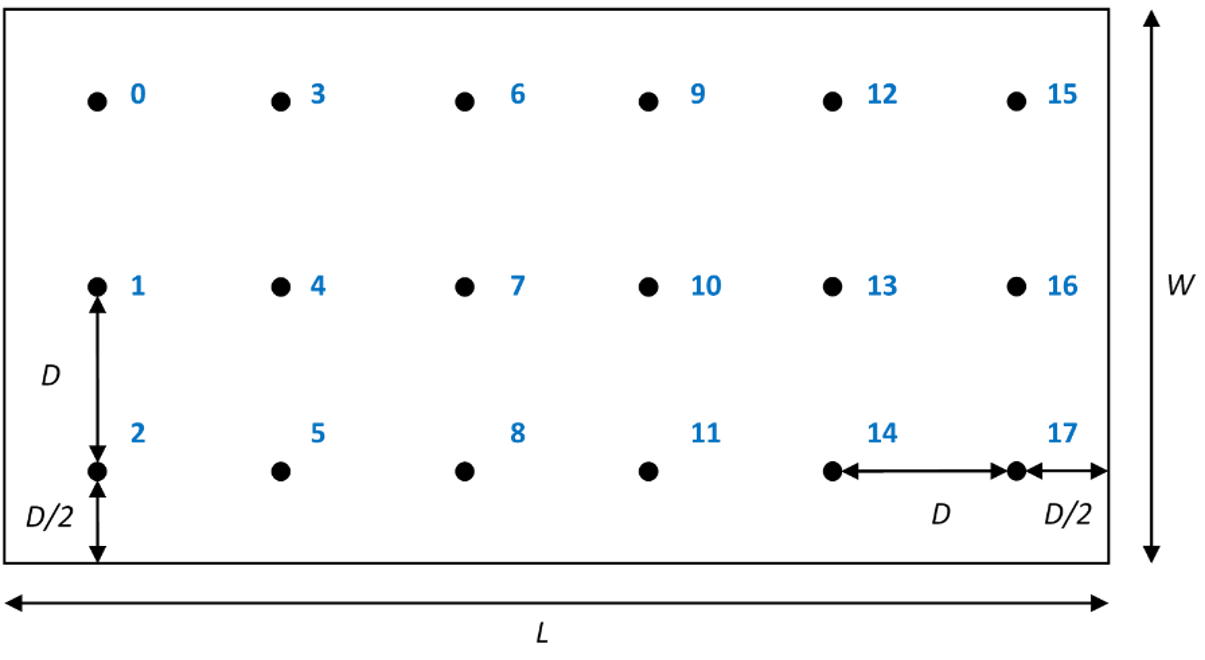}
	\centering
%\vspace{-0.3cm}
	\caption{3GPP Indoor Factory (InF) layout with $N_{\rm S}=18$ sensor nodes. For the InF-DH scenario, the dimensions are $L=120$~m, $W=60$~m and $D=20$~m, with 60\% of the area covered by clutters of 6m height and 2m size. }
	\label{Hall}
\vspace{-0.35cm}
\end{figure}

% \subsection{Description of Scenario and Datasets}
%The main scenario is a 120×60 m2 indoor factory with dense clutter and high base station height (InF-DH) scenario utilizing a 3.5 GHz carrier with 100 MHz signal bandwidth. 
We consider an indoor factory scenario with dense clutter and High sensor height (InF-DH) as in Fig.~\ref{Hall}, which illustrates the spatial layout of the 3GPP InF-DH environment, including the arrangement of the sensor nodes connected to a 5G base station, operating at $3.5$~GHz with a signal bandwidth of $100$~MHz. A total of $N_{\rm S}=18$ sensor nodes are uniformly distributed, spaced $20$~m apart, with the perimeter sensor nodes located $10$~m from the walls. The heights of the sensor nodes and devices are fixed at $8$~m and $1.5$~m, respectively. The spatial coordinates of the \(r\)-th sensor are represented as a \(\boldsymbol{\psi}_r = [x_{\rm s}^{r}, y_{\rm s}^{r}, z_{\rm s}^{r}]^\top \in \mathbb{R}^3\). The positions of all \(N_{\rm S}\) sensors are then collectively represented by  \(\boldsymbol{P} = [\boldsymbol{\psi}_1 \, \boldsymbol{\psi}_2 \, \cdots \, \boldsymbol{\psi}_{N_{\rm S}}] \in \mathbb{R}^{3 \times N_{\rm S}}\), where each column corresponds to the coordinates of a single sensor. A single transmit antenna port is assumed for the device. Each sensor node is equipped with a dual-polarized receive antenna,  enabling the reception of multi-antenna signals (MAS) through A=2 antenna ports. The RF channels between a device and the sensor nodes are generated using the 3GPP channel model documented in 3GPP Technical Report 38.901 \cite{3GPP_TR_38_901}.

% This figure 3GPP depicts the InF-DH scenario with $N_{\rm S}=18$ sensor nodes ($L=120$~m, $W=60$~m and $D=20$~m). With $60$\% area covered by clutters of $6$~m height, there is essentially no LoS link between a device and any of the %$N_{\rm S}=18$ distributed 
%     sensor nodes. 

% This figure itself depicts the 3GPP InF scenario, which focuses on factory halls of various sizes and varying levels of clutter density. Specifically, we study the InF-DH scenario with $N_{\rm S}=18$ sensor nodes, selecting a large factory hall with BSs at a height of 8m and a clutter configuration of {60\%, 6m, 2m}.

The 3GPP InF-DH environment features a variety of obstructions, such as small to medium-sized metallic equipment and irregularly shaped items like assembly and assorted machinery. In this paper, we focus on the InF-DH $\{60\%,~6~\text{m},~2~\text{m}\}$ configuration where $60$\% of the environment is covered by clutters of 6m height and 2m width/length. The LoS probability between a device and a sensor node is 0.008, based on empirical observations from the dataset for the InF-DH scenario \cite{3GPP_TR_38_901}. That is, a LoS link between a device and the distributed sensor nodes is very unlikely.
To locate a device, the device is configured to transmit the 5G sounding reference symbols (SRS) $\{ s_{\rm k} \}_{k=-N/2}^{N/2-1}$, where $N=3264$ is the number of SRS subcarriers, in the uplink to be received by the $18$ sensor nodes. The device transmit power is assumed $23$~dBm. The SRS signals have a comb spacing of $K_{\text{TC}} =2 $ subcarriers with $\Delta_f=30$ kHz subcarrier spacing and offset between $N_{\text{symb}}=2 $ consecutive OFDM symbols.
The signal being sampled by each sensor node at a rate of \( f_{\text{s}} = 4096 \cdot \Delta_{\text{f}} = 122.88 \) MHz is converted into the frequency domain (FD) using an FFT of size \( N_{\text{FFT}} = 4096 \). Assuming $L$  channel taps with complex gains  $\{ c_{a,l} \}_l$  and delays  $\{ \tau_l \}_l$ , the received SRS at the $(k + \frac{N}{2})$-th sub-carrier at receive antenna port $a$ is given by
\begin{equation}
\begin{aligned}
R_a[k] = s_k \sum_{l=0}^{L-1} c_{a,l} \exp(-j 2 \pi k \Delta_{ f} \tau_l) + W_a[  k]~,
\label{eq:SRS}
\end{aligned}
\end{equation}
where $W_a[k]$ is the received noise. The bandwidth of the SRS signal is $N\cdot \Delta_{f}=97.92$ MHz. The measured FD channel response (CR) is obtained as 
\begin{equation}
\begin{aligned}
H_a[k] \triangleq s_{k}^{\ast} R_a[ k]~.
\label{eq:CR}
\end{aligned}
\end{equation}
Assuming low mobility, the combed but offset FD CRs from the two consecutive OFDM symbols are coalesced into a combined FD CR for all subcarriers. The measured time domain (TD) channel impulse response (CIR) is obtained by further applying an inverse FFT:
\begin{equation}
\begin{aligned}
\{ h_a[d] \}_d \triangleq \text{IFFT}_{N_{\text{FFT}}} \left( \{ H_a[k] \}_{k=-\frac{N}{2}}^{\frac{N}{2}-1} \right)~.
\label{eq:TDCIR}
\end{aligned}
\end{equation}
where $d$ represents the discrete delay index. The TD power delay profile (PDP) is obtained by summing the powers over the antenna ports for each sample:
\begin{equation}
\begin{aligned}
p[d] \triangleq \sum_{a=0}^{A-1} |h_a[d]|^2~.
\label{eq:pdp}
\end{aligned}
\end{equation}
The PDP can be truncated to the first $N_{\text{ts}}=128$ time samples without losing much information about the link between a device and a sensor node. Note that we intentionally keep the processing of the received RF signals to the minimum and leave any signal processing (e.g., filtering, smoothing, or channel estimation) to the DL model. Unlike RSS or AoA-based features, which suffer from instability and degraded accuracy in NLOS conditions, the proposed approach utilizes the PDP as the input representation. While CSI provides richer frequency-domain information, it is complex-valued and high-dimensional, leading to substantially higher computational overhead. In contrast, the PDP offers a real-valued, compact, and stable time-domain representation of multipath propagation, making it well-suited for the lightweight and resource-efficient design.

With $N_{\rm S} = 18 $ sensor nodes, each device position can be considered to map uniquely to a PDP matrix $\mathbf{\Psi}$  of dimensions $N_{\rm S}\times N_{\textrm{ts}}$, with each row capturing the PDP at a different sensor. We considered three datasets ($\mathcal{D}$), namely small, medium, and large, sampling 1$0,000$, $20,000$, and $40,000$ uniformly randomly placed device locations, respectively, and recorded the corresponding PDPs. For the test datasets, we sample another $4,000$ uniformly randomly placed device locations and the corresponding TD PDPs. The PDPs collected from the $N_{\rm S}=18$ sensor nodes are used as inputs to the DL model (to be defined in Section \ref{section:DesignTransformer}) for joint processing, which will then output the 2D coordinates of the device.

\subsection{Problem Formulation}
We aim to learn a parametric model $f_{\boldsymbol{\Theta}}$ that estimates the 2D device location $\hat{\mathbf{y}} \in \mathbb{R}^2$ from the input PDPs $\boldsymbol{\Psi} \in \mathbb{R}^{N_{\mathrm{S}} \times N_{\mathrm{ts}}}$. The Transformer model parameters are denoted by $\boldsymbol{\Theta}$, and the predicted location is given by $\hat{\mathbf{y}} = f_{\boldsymbol{\Theta}}(\boldsymbol{\Psi})$, with the ground-truth location $\mathbf{y} \in \mathbb{R}^2$.

The learning objective is to minimize the expected 2D localization error, measured by the $L_2$ norm, while enforcing a computational constraint and ensuring robustness under data-limited conditions. This can be expressed as the following regularized optimization problem:
\begin{equation}
\begin{aligned}
\min_{\boldsymbol{\Theta}} \quad & 
\mathbb{E}_{(\boldsymbol{\Psi}, \mathbf{y}) \sim \mathcal{D}} 
\!\left[\left\| f_{\boldsymbol{\Theta}}(\boldsymbol{\Psi}) - \mathbf{y} \right\|_2^2\right]
+ \mathcal{R}_{\mathcal{D}}\!\left(f_{\boldsymbol{\Theta}}\right) \\
\text{s.t.} \quad & 
\mathrm{FLOPs}\!\left(f_{\boldsymbol{\Theta}}\right) \leq \mathcal{B}
\end{aligned}
\label{eq:optimization}
\end{equation}

The first term in \eqref{eq:optimization} represents the mean-squared localization loss, which directly drives the optimization of the model parameters $\boldsymbol{\Theta}$. The second term, $\mathcal{R}_{\mathcal{D}}\!\left(f_{\boldsymbol{\Theta}}\right)$, acts as a data-efficiency regularizer that enhances model robustness under limited training data. This robustness is further promoted through the proposed tokenization strategy, proposed Transformer-based model, and data augmentation, which together reduce dependence on large datasets and enable effective learning from diverse PDP realizations. The constraint on $\mathrm{FLOPs}\left(f_{\boldsymbol{\Theta}}\right)$ enforces a computational budget $\mathcal{B}$, ensuring that the total computational cost of the model does not exceed this limit. In this work, three model configurations are defined: small, medium, and large, corresponding to computational budgets of approximately $\mathcal{B}=4.5$M, $16.5$M, and $63.5$M FLOPs, respectively. The proposed architecture adheres to this constraint through a series of design optimizations, such as removing positional embeddings and the class token, adopting GAP, and carefully selecting the number of Transformer layers as well as the embedding and hidden dimensions.

\section{Proposed Framework for Preprocessing and Tokenization} \label{section:TokenPre}
 % The preprocessing framework ensures optimal signal conditioning, feature extraction, and preparation for subsequent analysis.
In this section, we present the proposed preprocessing technique designed for efficiently handling distributed MIMO sensor signals. Following this, we detail the conventional tokenization methods employed as benchmarks and introduce the proposed tokenization technique, which represents a significant advancement in the processing pipeline. This novel method is specifically tailored to address the limitations of existing approaches, thereby enhancing the representation and utilization of the sensor signal data.

\subsection{Pre-processing for RF Signals from Distributed Sensors}

\begin{figure}
%\vspace{-0.5cm}
	\includegraphics[width=0.99\linewidth]{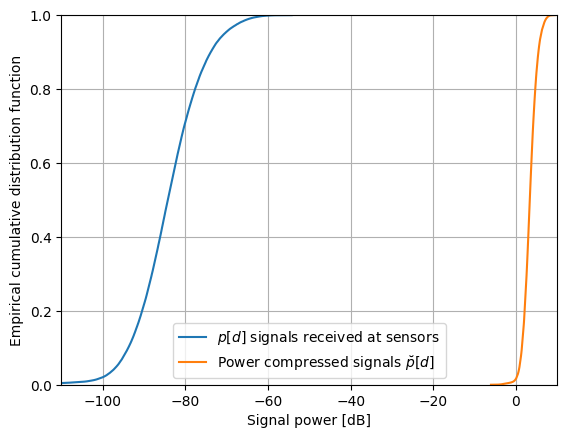}
	\centering
%\vspace{-0.3cm}
	\caption{Cumulative distribution function (CDF) of the received signal powers at different sensor nodes before and after power compression.}
	\label{PowerCompression}
\vspace{-0.35cm}
\end{figure} 

One major issue with processing distributed sensor network signals in a DL model is the wide dynamic range of the received signals. Because of path losses, reflection, shadowing, and fast fading effects, the received signal powers at different sensors can differ by over $35$~dB as illustrated by the cumulative distribution function (CDF) in Fig.~\ref{PowerCompression}. Without proper handling, a DL model will ignore weaker received signal streams. 

To address this dynamic range issue, we apply the power compression algorithm \cite{khamesi2024} on the PDP before feeding the signals to the DL model. For a length-$N_{\text{ts}}$ PDP $p[d]$, the power-compressed signal is given by: 
\begin{equation}
    \check{p}[d] = S^2 \| \{p[d]\} \|^{\frac{1}{r}-1} p[d]~,
\end{equation}
where $\| \{p[d]\} \| = \sum_{d=0}^{N_{\text{ts}}-1} p[d]$ is the total received power at a sensor over all its antenna ports, $r$ is the amplitude compression ratio parameter and $S$ is the target scale parameter after compression. For our indoor factory scenario in Fig.~\ref{Hall}, we use $r=5$ and $S=10$ to compress the received signal powers into a narrower range between $0$ and $10$~dB illustrated in Fig.~\ref{PowerCompression}. Furthermore, we found better performance is obtained when using square root PDP, $\sqrt{\check{p}[d]}$, instead of straight PDP.

\subsection{Conventional Tokenization Approaches}
%\subsection*{1) Conventional Tokenization Approaches:}

\textbf{\emph{Patch-based Tokenization (PBT)}}: In the wireless positioning literature, Transformer inputs are often represented in image format, with some approaches treating the CSI matrix as an image \cite{prasad2024vision}, and others converting RSSI and AoA data into images \cite{zhou2024vtil}, \cite{pan2022transformer}. For the first tokenization approach, similar to methods used in wireless literature, the PDP $\Psi$ which is a matrix of dimensions \(N_{\rm S} \times N_{\textrm{ts}}\), is treated as a 2D image. This image is partitioned into fixed-size patches (referred to as tokens) with dimensions \(W_{\rm h} \times W_{\rm w}\), where \(W_{\rm h}\) and \(W_{\rm w}\) denote the height and width of each token, respectively. Consequently, the total number of tokens generated from the PDP is given by \(N_{\rm tk} = \frac{N_{\rm S}}{W_{\rm h}} \cdot \frac{N_{\text{ts}}}{W_{\rm w}}\). Each token is then reshaped into a flattened vector, resulting in a spatial size (i.e., number of sample per token) of \(N_{\rm st} = W_{\rm h} \cdot W_{\rm w}\), this tokenization technique is referred to as \emph{Patch-based Tokenization (PBT)}. However, transforming inputs into image format and applying patching techniques fuses information from distinct physical measurements without considering the physical meaning of samples, which may hinder the Transformer from learning meaningful attention maps without effectively capturing the nuances of wireless communication environments. The PBT framework has not been previously utilized in conjunction with PDP in the literature. To the best of our knowledge, this paper is the first to explore its application with PDP.
\begin{figure}
%\vspace{-0.5cm}
\includegraphics[width=0.85\linewidth]{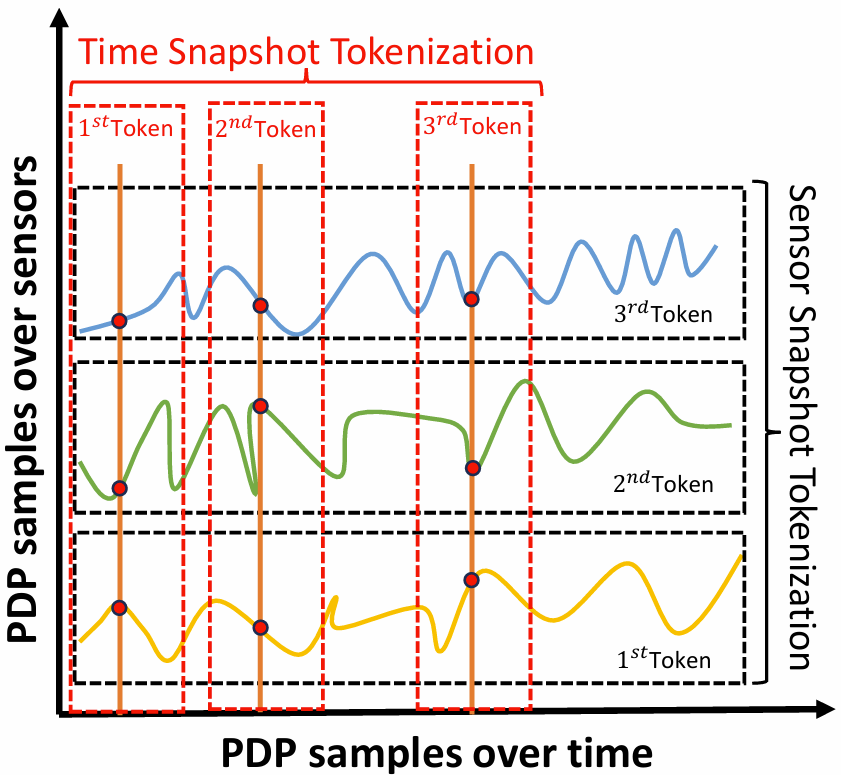}
	\centering
%\vspace{-0.3cm}
	\caption{Example of PDP tokenization techniques for  $N_{\rm S}=3$ sensors.
    The proposed \emph{Sensor Snapshot Tokenization (SST)} treats each sensor\textquotesingle s PDP vector of size $1\times N_{\rm ts}$ as a token, while the alternative \emph{Time Snapshot Tokenization (TST)} uses vectors of time samples across all sensors at a single time instant as tokens. 
    }
	\label{Tokenization}
\vspace{-0.35cm}
\end{figure} 
% as an example to demonstrate the tokenization techniques. Proposed tokenization technique, \emph{Sensor Snapshot Tokenization (SST)}, where a PDP vector of size $1\times N_{\rm tap}$ from each $N_{\rm S}$ sensors is a token. An alternative tokenization technique, namely \emph{Time Snapshot Tokenization (TST)}, is also shown, where the vector of time samples across all sensors at a time instant constitutes a token.

% \textcolor{red}{Considering this we reflect on the Transformer's poor performance and propose an efficient way for tokenization which will result in meaningful attention maps and will lead to a low-complexity Transformer model. }
% To address the limitations of the previously discussed tokenization methods, we introduce a novel

\subsection{Proposed Tokenization Approaches}
%\subsection*{2) Proposed Tokenization Approach:}
%\textbf{\emph{Sensor Snapshot Tokenization (SST)}}:
In this section, we introduce two novel tokenization approaches\footnote{Both SST and TST can be considered as special cases of \emph{PBT}; however, these specific approaches have not been studied in the literature.}—\emph{Time Snapshot Tokenization (TST)} and \emph{Sensor Snapshot Tokenization (SST)}. While both methods are designed to enhance feature representation in wireless communication systems, SST is the primary focus of this work due to its superior ability to capture spatial features and address the limitations of traditional input formulations. TST is presented as a complementary approach to highlight the strengths of SST.
% The second tokenization technique, referred to as \emph{Time Snapshot Tokenization (TST)}, considers the PDP value from all \(N_{\rm S}\) sensors at a single time step as a single token, resulting in a total of \(N_{\rm tk} = N_{\text{tap}}\) tokens, each with a dimensionality of \(N_{\rm st} = N_{\rm S}\), effectively embedding multivariate temporal information within each token.

\textbf{\emph{Time Snapshot Tokenization (TST)}}: We have \( N_{\rm S} = 18 \) sensors; for illustration, assume that the representation of the PDP from 3 sensors is depicted in Fig.~\ref{Tokenization}. The second tokenization technique, referred to as \emph{Time Snapshot Tokenization (TST)}, represents the PDP values from all \(N_{\rm S}\) sensors at a single time step as a single token, resulting in 
\(N_{\rm tk} = N_{\text{ts}}\) tokens, each with a dimensionality of \(N_{\rm st} = N_{\rm S}\), effectively embedding multivariate temporal information within each token. When these temporal tokens are input into the multi-head attention ($\mathsf{MHA}$) (to be defined in Section \ref{section:DesignTransformer}) mechanism, the model tends to prioritize numerical values over the semantic relationships inherent among samples taken at the same time. Furthermore, since the values in each token are derived from different sensors, they may embody entirely distinct meanings, leading to the loss of multivariate correlations.

Tokens generated from this tokenization technique suffer from excessively localized receptive fields and will not be able to convey meaningful information, as each token is based on a single time sample across $N_{\rm S}$ sensors. Given that variations in time series data are heavily influenced by the order of the sequence, the permutation-invariant nature of the $\mathsf{MHA}$ mechanism is ill-suited for such data structures. As a result, the Transformer's ability to capture essential time series representations and portray multivariate correlations is diminished, thereby limiting its capacity and generalization ability across diverse time series datasets.

\textbf{\emph{Sensor Snapshot Tokenization (SST)}} \label{SST}: Considering the limitations of the previously discussed tokenizations we reflect on the Transformer's poor performance and propose an efficient way for tokenization technique called \emph{Sensor Snapshot Tokenization (SST)}. This approach independently embeds each PDP series received by a sensor into a distinct token, as illustrated by the black dashed boxes in Fig.~\ref{Tokenization}. Where the number of tokens is given by \( N_{\rm tk} = N_{\rm S} \), with each token having a dimensionality of \( N_{\rm st} =N_{\textrm{ts}} \). This independent embedding enhances the receptive field, enabling the Transformer model to learn meaningful representations. Importantly, this tokenization strategy acknowledges that the information from each sensor operates independently, reflecting channel independence between sensors. Each token encapsulates a unique physical meaning, allowing Transformer to focus more on the semantic relationships among time samples rather than their numerical values. 

%Understanding these correlations is crucial when choosing tokenization methods for multivariate time series forecasting.
The resulting tokens represent the global characteristics of the series, producing variate-centric tokens, and this approach allows the $\mathsf{MHA}$ to cross-query and correlate received signals from all sensor nodes in parallel to capture multi-variate correlations. This is aligned with the conventional signal processing principle of processing signals from all sensors/antennas jointly. Capturing multi-variate correlation relationships enhances the interpretability of our models, especially since the spatial arrangement of the sensors remains consistent throughout, these correlations reflect the relative positions and interactions between sensors, which can provide valuable clues about the location of the signal source. These factors will decrease the heavy dependence of Transformer on large datasets. Additionally, an $\mathsf{MLP}$ will effectively derive generalizable representations from these tokens.

The second critical aspect of the \emph{SST} technique is its potential to achieve reduced computational complexity. The $\mathsf{MHA}$ layer, the most computationally heavy block of a Transformer, exhibits quadratic complexity with respect to the number of tokens. In the first approach, \emph{PBT}, the number of tokens depends on the parameters \(W_{\rm h}\) and \(W_{\rm w}\); smaller parameter values result in a higher number of tokens. Larger values lead to less number of tokens but will aggregate information across multiple physical measurements. This aggregation can dilute variable-specific representations and lead to less informative attention maps. In contrast, the second approach \emph{TST} maintains a token count of \(N_{\rm tk} = N_{\text{ts}}\), while the third approach \emph{SST} has \(N_{\rm tk} = N_{\rm S}\). Given that \( N_{\rm S} \ll N_{\text{ts}} \), the complexity of the third approach is significantly lower, resulting in greater computational efficiency.

% \textcolor{green}{The matrix \( QK^T \) illustrates the correlations between tokens, where each entry \( (i, j) \) is derived from the product of the query vector \( q_i \) and the key vector \( k_j \). This creates a multivariate correlation matrix that, when multiplied by the value vectors \( V \), highlights those tokens that share strong correlations. When two tokens demonstrate a strong correlation, it suggests that their TD-PDPs are similar. This similarity implies that the underlying device is closely aligned along the boundaries defined by these sensors.}
\begin{figure*}[t] % Use figure* to span both columns
    \centering
    \subfloat[\emph{Vanilla-T} model.]{%
        \includegraphics[width=0.43\textwidth]{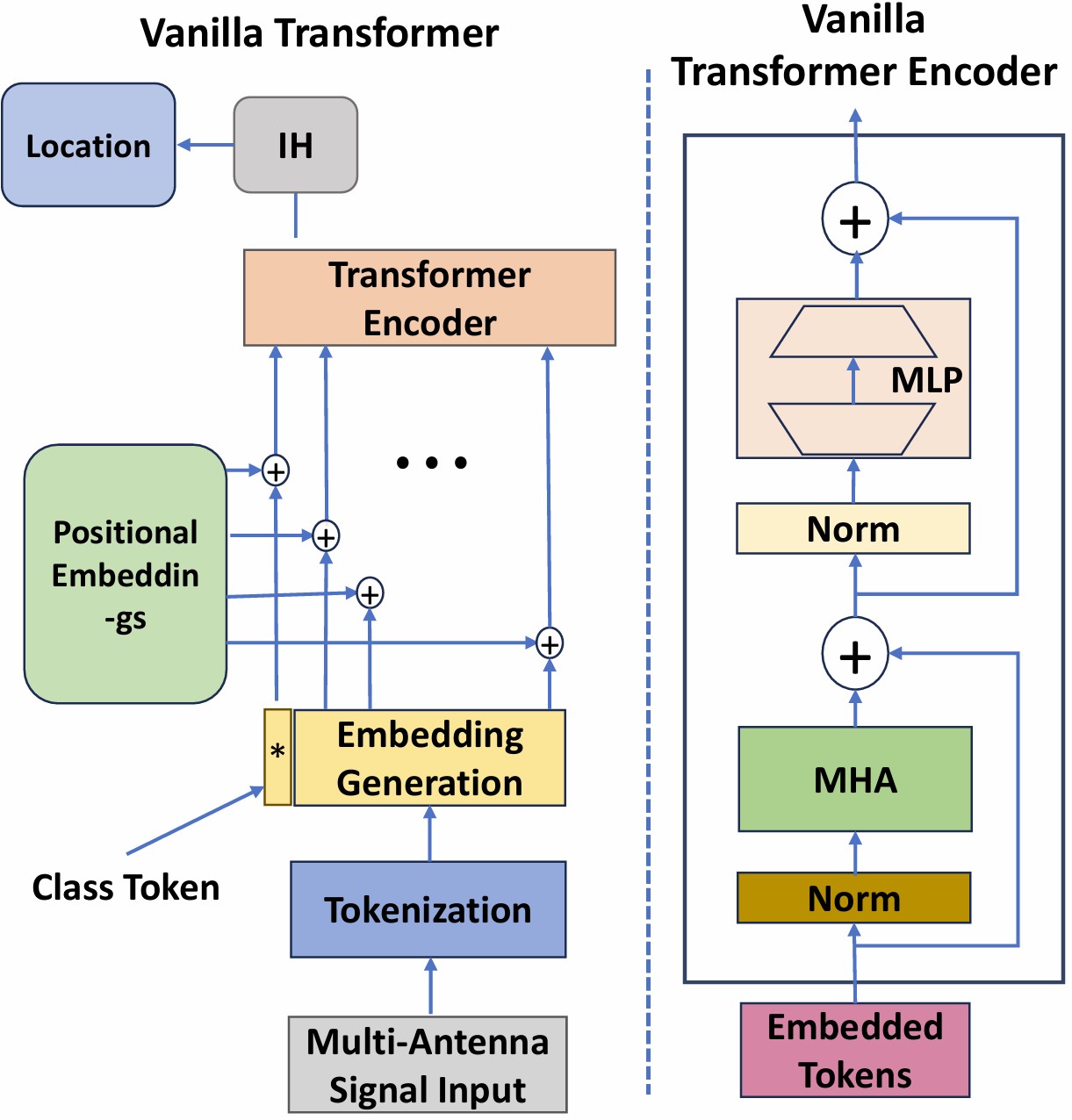}
        \label{fig:ViT1}}
    \quad % or \hspace{0.05\textwidth}
    \subfloat[Lightweight Swish-Gated Transformer model (\emph{L-SwiGLU-T}).]{%
        \includegraphics[width=0.43\textwidth]{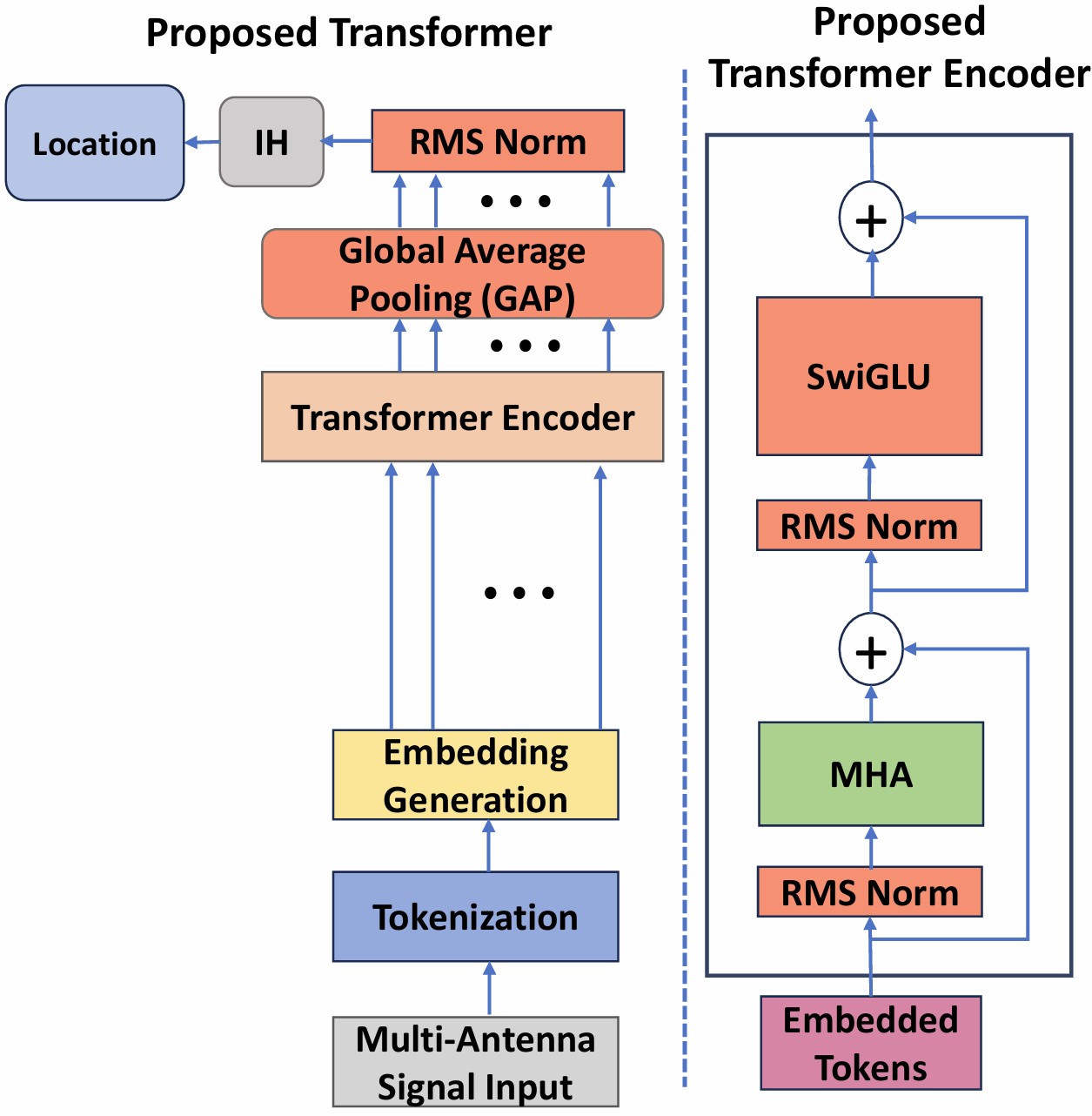}
        \label{fig:ViT2}}
    \caption{The architecture of the \emph{vanilla-T} model, an encoder-only Transformer, is depicted in Fig. \ref{fig:ViT1}, while the proposed lightweight Swish-Gated Transformer, with the modified parts highlighted in light red, is shown in Fig. \ref{fig:ViT2}.}
    \label{fig:ViT_Model}
    \vspace{-5mm} % Adjust this value if needed
\end{figure*}
\section{Designing Transformer Model for Distributed Sensor Networks}  \label{section:DesignTransformer}
% With the recent success of encoder-only Transformer architectures like ViT in computer vision, there has been growing interest in adapting them for indoor positioning. However, directly applying ViT to this domain results in suboptimal performance and significantly increases model complexity. 

In this work, we design an encoder-only Transformer architecture tailored for device localization in cluttered indoor environments dominated by severe NLoS conditions. Specifically, the Transformer is designed to learn the complex mapping $f_{\boldsymbol{\Theta}}$ between the device's position $y$ and the PDP $\mathbf{\Psi}$ received from all sensor nodes $N_{\rm S}$.
The PDP, obtained from channel measurements at the sensors, encodes the multipath propagation characteristics by capturing the power and delay of reflected and scattered signals. This information inherently represents the spatial geometry and structure of the environment, enabling the Transformer to accurately learn the relationship between the PDP and the device’s position \(\Ytb\).

In this section, we describe the architectures of the baseline \emph{Vanilla Transformer (Vanilla-T)} and the proposed \emph{Lightweight Swish-Gated Linear Unit Transformer (L-SwiGLU-T)}. The complete model structures are illustrated in Fig.~\ref{fig:ViT_Model}.

% In this section, we describe the architecture of the \emph{Vanilla Transformer (Vanilla-T)} model (i.e., ViT) and the proposed lightweight Swish-Gated Linear Unit-based Transformer model, referred to as \emph{L-SwiGLU-T}. The complete architecture of \emph{vanilla-T} and \emph{L-SwiGLU-T}  model can be seen in the Fig.~\ref{fig:ViT_Model}. 

% The vanilla ViT utilizes a standard encoder-only Transformer structure, incorporating a class token for predictions. In contrast, the proposed \emph{L-SwiGLU ViT} eliminates the use of the class token and replaces the conventional multi-layer perceptron (MLP) with Swish-Gated Linear Units (SwiGLU) as can be seen in Fig. \ref{fig:ViT_Model}, enhancing performance and computational efficiency. Additional architectural differences between the two models are detailed below.

% \begin{figure*}[t] % Use figure* to span both columns
%     \centering
%     \includegraphics[width=1\textwidth]{COMP_ViT.jpg} % Adjust to span the text width
%     \caption{The architecture of the vanilla Transformer model, an encoder-only Transformer, is depicted on the left, while the proposed lightweight Swish-Gated Transformer, with the modified parts highlighted in red, is shown on the right.}
%     \label{fig:ViT_Model}
%     \vspace{-0.35cm} % Adjust spacing if needed
% \end{figure*}
\vspace{-3mm}
\subsection{Vanilla Transformer (Vanilla-T) Architecture}
Fig.~\ref{fig:ViT1} illustrates the architecture of the \emph{Vanilla-T} model. The localization process using a Transformer can be categorized into $3$ main steps: $1$) Input Embedding Preparation $2$) Encoder Mechanism, and $3$) Position Estimation.

% \subsection*{$1$) Input Embedding Preparation:}
\subsubsection{\textbf{Input Embedding Preparation}}\label{Inputembed}
In the embedding preparation, the PDP input is first converted into tokens/patches $\mathbf{X}_{\text{PDP}} \in \mathbb{R}^{N_{\rm tk} \times N_{\rm st}}$ using the tokenization method, where $N_{\rm tk}$ is the number of tokens generated from PDP input, and $N_{\rm st}$ is the number of samples in each token. After this, the PDP tokens $\mathbf{X}_{\text{PDP}}$ are linearly projected to create embeddings from the input using a learnable projection, $\mathbf{E} \in \mathbb{R}^{N_{\rm st} \times D_{\text{emb}}}$  resulting in \( \mathbf{X}_{\text{inp}} = \mathbf{X}_{\text{PDP}} \mathbf{E} \in \mathbb{R}^{N_{\rm tk} \times D_{\text{emb}}} \) where $D_{\text{emb}}$ is the dimensionality of the embeddings. Embeddings are a way to represent data in a continuous vector space, where each PDP token is mapped to a dense, one-dimensional representation. This process captures the essential features of the PDP tokens, making it easier for the model to learn patterns and make predictions. 

%  \begin{equation}
% \begin{aligned}
% X_{\text{emb}} &= X_{\text{MAS}} E \quad \in \mathbb{R}^{T_t, D_{\text{emb}}}
% \end{aligned}
% \label{eq:emb}
% \end{equation}

The next step involves appending a learnable embedding, referred to as the class token \(\mathbf{X}_{\text{cls}} \in \mathbb{R}^{1 \times D_{\text{emb}}}\), to the sequence of embeddings generated from the input \(\mathbf{X}_{\text{inp}}\). This class token serves as a global representation for the model. A classification head will be attached to it for making predictions. The resulting embedding sequence is represented as \(\mathbf{X}_{\text{emb}} = [\mathbf{X}_{\text{inp}}, \mathbf{X}_{\text{cls}}] \in \mathbb{R}^{\tilde{N}_{\rm tk} \times D_{\text{emb}}}\).  The total number of tokens in embedding $\mathbf{X}_{\text{emb}}$ will be $\tilde{N}_{\rm tk} =N_{\rm tk}+1$, where 1 comes from the additional class token $\mathbf{X}_{\text{cls}}$. Embeddings are further encoded by positional embeddings resulting in:
  \begin{equation}
 \Ztb=\mathbf{X}_{\text{emb}}+\mathbf{E}_{\text{pos}}~,
\label{eq:PosE}
\end{equation}
 where $\mathbf{E}_{\text{pos}}\in \mathbb{R}^{\tilde{N}_{\rm tk} \times D_{\text{emb}}}$ are learnable parameters that capture positional information. 

\begin{figure*}[h!]
%\vspace{-0.5cm}
	\includegraphics[width=0.8\linewidth]{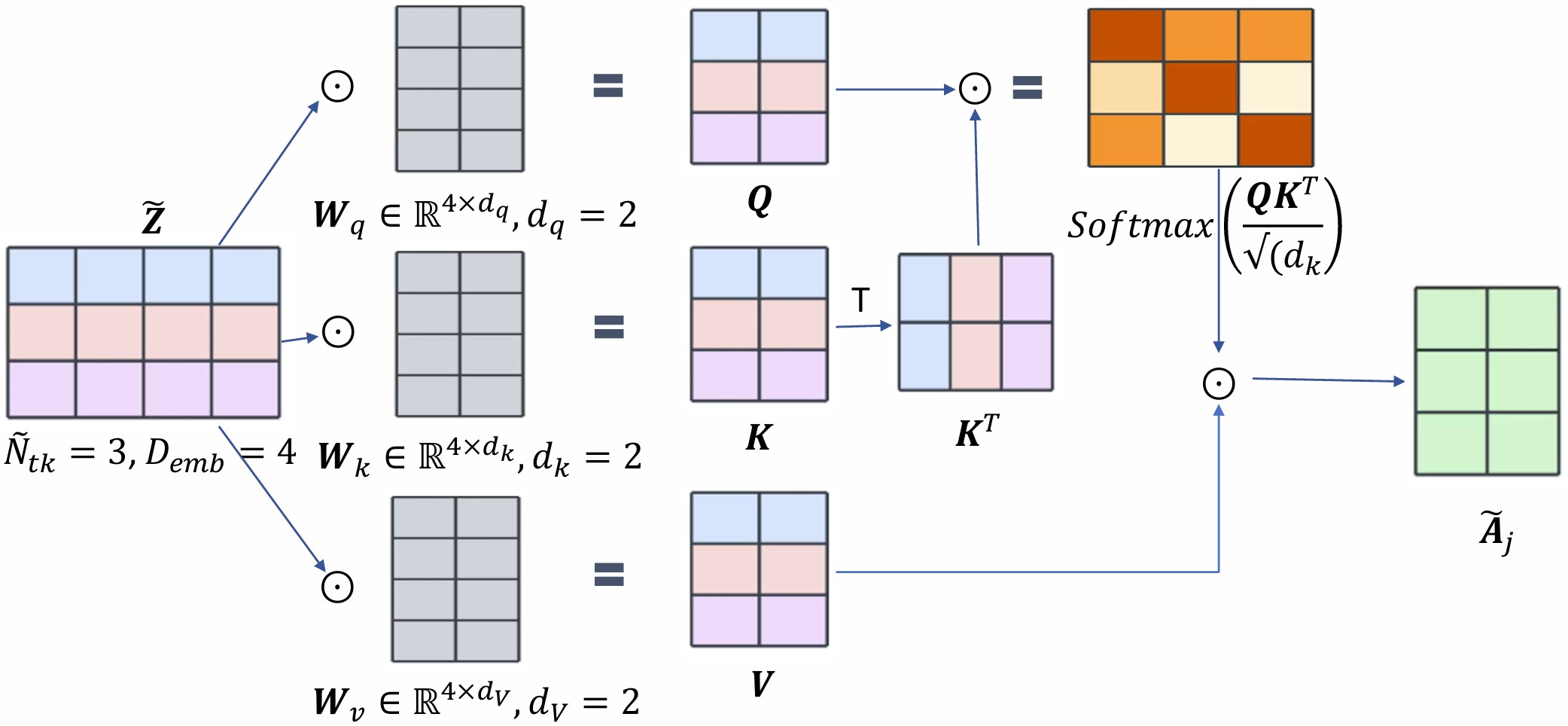}
	\centering
%\vspace{-0.3cm}
	\caption{ Scaled dot-product attention mechanism with \( \tilde{N}_{\rm tk} = 3 \) tokens and embedding dimension \( D_{\text{emb}} = 4 \). The input \(\tilde{\mathbf{Z}}\) is projected into \(\mathbf{Q}\), \(\mathbf{K}\), and \(\mathbf{V}\) using weight matrices \(\mathbf{W}_q\), \(\mathbf{W}_k\), and \(\mathbf{W}_v\), and attention is computed using the scaled dot product of \(\mathbf{Q}\) and \(\mathbf{K}^T\).}
	\label{Atten}
\vspace{-0.35cm}
\end{figure*} 

% \begin{equation}
% \resizebox{\columnwidth}{!}{%
%     $Q = \Ztb W_q, \; K = \Ztb W_k, \; V = \Ztb W_v \quad Q \in \mathbb{R}^{N_t \times d_q}, \;\\ K \in \mathbb{R}^{N_t \times d_k}, \; V \in \mathbb{R}^{N_t \times d_v}$
% }
% \label{eq:Proj}
% \end{equation}

% \subsection*{2) Encoder Mechanism:}
\subsubsection{\textbf{Encoder Mechanism}} \label{EncoderMech}
The encoded PDP tokens $\Ztb$ are subsequently processed by the $N_{\rm L}$ Transformer encoder blocks, with each encoder block comprised of four primary components: the Multi-Head Attention ($\mathsf{MHA}$) block, the $\mathsf{MLP}$, residual connections, and a normalization layer. The main component that distinguishes the Transformer from other DL models is the $\mathsf{MHA}$ mechanism. In this mechanism, the embeddings interact to exchange and share information. Within the attention block, communication is facilitated by projecting the encoded embedding matrix $\Ztb$ to generate the Queries ($\mathbf{Q}$), Keys ($\mathbf{K}$), and Values ($\mathbf{V}$) matrices. This is achieved using learnable weight matrices $\mathbf{W}_q \in \mathbb{R}^{D_\text{emb}\times d_q}$, $\mathbf{W}_k \in \mathbb{R}^{D_\text{emb}\times d_k}$, and $\mathbf{W}_v \in \mathbb{R}^{D_\text{emb}\times d_v}$, respectively, as follows:
%Communication within the attention block involves projecting the encoded embedding matrix $\Ztb$ to generate Queries ($\mathbf{Q}$), Keys ($\mathbf{K}$), and Values ($\mathbf{V}$) matrix using learnable weight matrices of $\mathbf{Q}$, $\mathbf{K}$ and $\mathbf{V}$, represented by $\mathbf{W}_q \in \mathbb{R}^{D_\text{emb}\times d_q}$, $\mathbf{W}_k \in \mathbb{R}^{D_\text{emb}\times d_k}$ and $\mathbf{W}_v \in \mathbb{R}^{D_\text{emb}\times d_v}$ respectively as follows:
\begin{equation}
\begin{aligned}
\mathbf{Q} &= \Ztb \mathbf{W}_q, \quad \mathbf{K} = \Ztb \mathbf{W}_k, \quad \mathbf{V} = \Ztb \mathbf{W}_v, \\
\mathbf{Q} &\in \mathbb{R}^{\tilde{N}_{\rm tk} \times d_q}, \quad  \mathbf{K} \in \mathbb{R}^{\tilde{N}_{\rm tk}\times d_k},\quad \mathbf{V} \in \mathbb{R}^{\tilde{N}_{\rm tk} \times d_v}~.
\end{aligned}
\label{eq:Proj}
\end{equation}

The attention mechanism (Fig.~\ref{Atten}) is applied to  (\(\mathbf{Q}\)), (\(\mathbf{K}\)), and (\(\mathbf{V}\)) matrices to compute the correlation matrix \(\mathbf{C} \in \mathbb{R}^{\tilde{N}_{\rm tk}\times \tilde{N}_{\rm tk}}\), defined as:
 \begin{equation}
\begin{aligned}
\mathbf{C}=\left(\frac{\mathbf{Q} \mathbf{K}^T}{\sqrt{d_k}}\right)~,
 \end{aligned}
\label{eq:corellation}
\end{equation}
where \(d_k = D_{\text{emb}}\) serves as the normalization term. The \(\mathbf{C}\) quantifies the similarity between tokens, with each element representing the degree of correlation between corresponding token pairs. Then, the softmax is applied along the rows to turn similarity values into probabilities.

The attention score \(\mathbf{A}_j\) in the \( j^{\text{th}} \) Transformer encoder block, where \( j \in \{1, 2, \cdots, N_{L}\} \) and \( N_{L} \) is the total number of Transformer encoder blocks, is computed as:
 \begin{equation}
\begin{aligned}
\Atb_j(\mathbf{Q}, \mathbf{K}, \mathbf{V}) = \mathsf{Softmax}(\mathbf{C})\mathbf{V}=\mathsf{Softmax}\left(\frac{\mathbf{Q} \mathbf{K}^T}{\sqrt{d_k}}\right)\mathbf{V}~,
 \end{aligned}
\label{eq:Attn}
\end{equation}
where \(j\in\{1, 2, \cdots, N_L\}\).
This operation assigns a weighted sum of the values (\(\mathbf{V}\)) to each embedding, where the weights are determined by the similarity between the query (\(\mathbf{Q}\)) and key (\(\mathbf{K}\)) vectors. The similarity calculation enables the model to learn relationships between tokens by distributing attention based on relevance, rather than solely relying on exact matches between \(\mathbf{Q}\) and \(\mathbf{K}\).

  % After the multiplication with V, each embedding gets a weighted sum of the values of other embeddings based on their Q, and K similarity. Calculating the similarity between \( Q \) and \( K \) is crucial because if values are assigned solely based on exact matches between the query and the key, there will be no learning.

In \cite{dosovitskiy2020image}, the authors hypothesized that instead of projecting the embeddings to generate $\mathbf{Q}$, $\mathbf{K}$ and $\mathbf{V}$ one time, it is more beneficial to project it $N_{\rm h}$ times using $N_{\rm h}$ different learnable weight matrices $\mathbf{W}^i_q$, $\mathbf{W}^i_k$, and $\mathbf{W}^i_v$, $i\in (1, \cdots, N_{\rm h})$, where $N_{\rm h}$ denotes the total number of heads. In this way, the model can learn $N_{\rm h}$ different representations from the input embeddings. To maintain the computational complexity of the multi-head attention, block similar to that of the single-head block, one can choose $d_q^i=d_k/N_{\rm h}$, with the same selection for $d_k^i$ and $d_v^i$. The overall complexity of $\mathsf{MHA}$ attention layer is $O(\tilde{N}_{\rm tk}^2 D_{\text{emb}})$.

After the attention for all heads is calculated, the output is concatenated into a single matrix with the help of the projection weight matrix \( \mathbf{W}_o \in \mathbb{R}^{N_{h}d_v\times D_{\text{emb}}} \), as follows:
\begin{gather}
\boldsymbol{\mathcal{H}}_j^i = \Atb_j\left( \mathbf{Q} = \Ztb \mathbf{W}_q^i, \, \mathbf{K} = \Ztb \mathbf{W}_k^i, \, \mathbf{V} = \Ztb \mathbf{W}_v^i \right), \nonumber\\
\mathsf{MHA}(\Ztb) = \mathsf{Concat} \left( \boldsymbol{\mathcal{H}}_j^1, \ldots, \boldsymbol{\mathcal{H}}_j^{N_{\rm h}} \right) \mathbf{W}_o.\nonumber
\end{gather}

% %\begin{equation}
% \begin{align}
% h_l^i &= \Atb_l\left( \mathbf{Q} = \Ztb \mathbf{W}_q^i, \, \mathbf{K} = \Ztb \mathbf{W}_k^i, \, \mathbf{V} = \Ztb \mathbf{W}_v^i \right) ~,\nonumber\\
%  \textrm{MHA}(\Ztb_l)&= \textrm{Concat} \left( h_l^1, \ldots, h_l^H \right) \mathbf{W}_o ~.\nonumber%\label{eq:MHAttn2}
% \end{align}

Before entering every block, the input is normalized using a LayerNorm ($\mathsf{LN}$), and the output is then added to the input through a residual connection to help with gradient flow and optimization:
\begin{equation}
\begin{aligned}
\Ztb_j^{\mathsf{MHA}} = \mathsf{MHA}\left(\mathsf{LN}\left(\Ztb_{j-1}\right)\right) + \Ztb_{j-1}  ~.
\end{aligned}
\label{eq:MHAttn}
\end{equation}
The $\mathsf{LN}$ operation for the input vector \( \Ztb_{l-1} \) is defined as:
\[
\mathsf{LN}(\Ztb_{j-1}) = \gamma \odot \left( \frac{\Ztb_{ j-1} - \mu}{\sqrt{\sigma_{\rm LN}^2 + \epsilon}} \right) + \beta~,
\]
where \( \mu \) and \( \sigma_{\rm LN}^2 \) represent the mean and variance of \( \Ztb_{l-1} \), respectively, and \( \gamma \) and \( \beta \) are learnable parameters.

The next key component of a Transformer is the $\mathsf{MLP}$ block, which comprises two layers. This block is responsible for learning the information encoded in each embedding. The first layer maps the input to a higher dimension called hidden dimension $h_{\text{dim}}$, and the second and last layer maps it to the $D_{\text{emb}}$ again. Therefore, the input and output of the Transformer block will have a similar shape. 
%\begin{equation}
\begin{align} (\Ztb_j^{\mathsf{MHA}})&=\mathbf{W}^2_j\left(\sigma\left(\mathbf{W}^1_j \Ztb_j^{\mathsf{MHA}} +b^1_j\right)\right)+b^2_j~,
\end{align}
% \label{eq:MHAttn}
%\end{equation}
where $\sigma$ is activation function and $\mathbf{W}^1_j,\mathbf{W}^2_j,b^1_j,$ and $b^2_j$ are weight and biases of $\mathsf{MLP}$ layers in $j^{\rm th}$ Transformer block.

Similar to the $\mathsf{MHA}$ block, for output of $j^{\rm th}$ encoder block an $\mathsf{LN}$ is applied before the $\mathsf{MLP}$ block, followed by a residual connection:
\begin{align}
\Ztb_j^{\mathsf{ENC}} &= \mathsf{MLP}_j \left(\mathsf{LN}\left(\Ztb_j^{\mathsf{MHA}}\right)\right)+\Ztb_j^{\mathsf{MHA}}~.   
\end{align}
\subsubsection{\textbf{Position Estimation}} \label{PositionEst}
The first embedding, referred to as the class token (\(\mathbf{X}_{\text{cls}}\)), is designed to capture a global representation by interacting with all other tokens through the attention mechanism. After processing through the \(N_{\rm L}^{\text{th}}\) Transformer encoder block, the class token (\(\mathbf{X}_{\text{cls}}\)) is forwarded to an inference head ($\mathsf{IH}$) which consist of one fully connected layer to generate the final predictions \(\Ytb\):
\begin{equation}
\begin{aligned}
\Ytb=\mathsf{IH}(\Ztb_{N_L}^{\text{ENC}}(0))~.
\end{aligned}
 \label{eq:pred}
\end{equation}
\subsection{Proposed \emph{L-SwiGLU ViT}}
This subsection provides a comprehensive step-by-step explanation of the modifications made to the Transformer components. The Fig.~\ref{fig:ViT2} depicts the architecture of the proposed \emph{L-SwiGLU-T} model, highlighting the introduced enhancements. The modifications made in the proposed model are categorized into two parts: $1$) Modifications Within the Transformer Encoder Block, and $2$) Modifications Outside the Transformer Encoder Block. These modifications are detailed in the following subsections.

\subsubsection{\textbf{Modifications Within the Transformer Encoder Block}}
To enhance the stability, robustness, and overall performance of the \emph{L-SwiGLU-T} model, the standard $\mathsf{LN}$ used in the vanilla-T model is replaced with a Root Mean Square (RMS) normalization layer.
LayerNorm performs both re-centering and re-scaling, making it robust to shift noise in inputs and weights. However, in line with the hypothesis proposed in \cite{zhang2019root}—which suggests that the success of $\mathsf{LN}$ is attributed primarily to its re-scaling invariance rather than re-centering invariance—we substitute $\mathsf{LN}$ with $\mathsf{RMSNorm}$. $\mathsf{RMSNorm}$ exclusively preserves re-scaling invariance while omitting re-centering, aligning with this hypothesis to further optimize the model's performance. The $\mathsf{RMSNorm}$ operation for the input \( \Ztb_{\rm j-1} \) is defined as:
\begin{equation}
\begin{split}
    \mathsf{RMSNorm} (\Ztb_{\rm j-1}) &= \frac{\Ztb_{\rm j-1}}{\mathsf{RMS}(\Ztb_{\rm j-1})} \odot \gamma, \\
    \text{where} \quad \mathsf{RMS} (\Ztb_{\rm j-1}) &= \sqrt{\frac{1}{n} \sum_{m=1}^{n} \left(\Ztb_{\rm j-1}^{(m)}\right)^2}~.
\end{split}
\end{equation}
Here, \( \gamma \) is a learnable scaling parameter, and $\mathsf{RMSNorm}$ avoids the mean subtraction step to simplify computation.

\begin{figure}
%\vspace{-0.5cm}
	\includegraphics[width=0.9\linewidth]{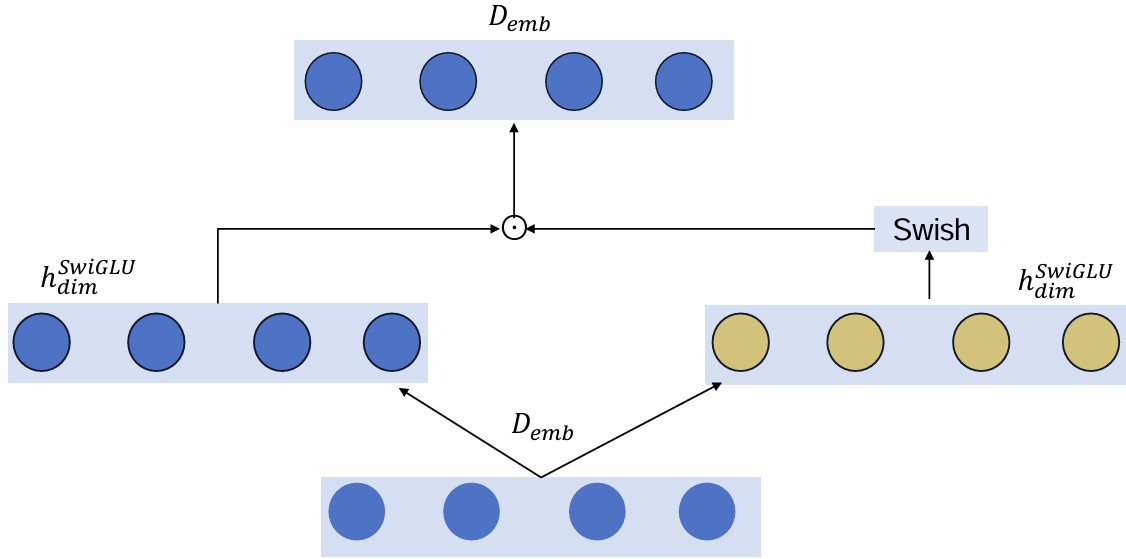}
	\centering
%\vspace{-0.3cm}
	\caption{GLU mechanism: one linear projection is gated via Swish activation, while the other bypasses activation; their element-wise product ($\odot$) filters relevant features.}
	\label{SwiGLU}
\vspace{-0.35cm}
\end{figure} 

In the proposed \emph{L-SwiGLU-T} model, the standard $\mathsf{MLP}$ is replaced with a Swish-Gated Linear Unit (SwiGLU)-based $\mathsf{MLP}$ to enhance performance \cite{shazeer2020glu}. Recognizing that not all information within each token contributes equally to predicting the location, we incorporated the Gated Linear Unit (GLU) within the $\mathsf{MLP}$ block to enable the model to focus on the most relevant features. Acting as an effective denoiser, the GLU is particularly beneficial in wireless communication scenarios where filtering out irrelevant data is critical. The GLU achieves this by performing a component-wise product of two linear projections of size $h^{\text{SwiGLU}}_{\text{dim}}$, selectively passing meaningful information.
\begin{equation}
\begin{aligned}
\mathsf{SwiGLU}_j (\Ztb_j^{\mathsf{MHA}}) =
    & \left(\mathsf{Swish}(\Ztb_j^{\mathsf{MHA}} \mathbf{W}_j^1) \odot \right. \\
    & \left. (\Ztb_j^{\mathsf{MHA}} \mathbf{W}_j^2) \right) \mathbf{W}_j^3~.
\end{aligned}
\label{eq:SwiGLU}
\end{equation}
The modified $\mathsf{MLP}$ block, referred to as $\mathsf{SwiGLU}$, processes the input \( \Ztb_j^{\mathsf{MHA}} \) through a gating mechanism, as illustrated in Fig.~\ref{SwiGLU}. Specifically, the input undergoes two linear projections, \( \Ztb_j^{\mathsf{MHA}} \mathbf{W}_j^1 \) and \( \Ztb_j^{\mathsf{MHA}} \mathbf{W}_j^2 \). One of these projections, \( \Ztb_j^{\mathsf{MHA}} \mathbf{W}_j^1 \), is passed through the Swish activation function, which enables non-linear transformations essential for capturing complex patterns. The activated output is then combined with the second projection via a Hadamard product (\( \odot \)), creating a selective gating mechanism that prioritizes relevant features while suppressing noise. The resulting representation is projected through a final linear transformation \( \mathbf{W}_j^3 \).
% , enhancing the expressiveness of the MLP block and improving the model’s ability to handle complex data dependencies. This modified structure, SwiGLU, significantly boosts model performance by refining feature selection and representation.

% The output of the SwiGLU block is represented as:
The output of the 
$j^{\rm th}$ encoder block is computed as:
\begin{equation} \label{eq:SWIGLU}
\Ztb_j^{\text{ENC}} = \mathsf{SwiGLU}_j\left(\mathsf{RMSNorm}\left(\Ztb_j^{\mathsf{MHA}}\right)\right) + \Ztb_j^{\mathsf{MHA}}~,
\end{equation}
where the output of the $\mathsf{MHA}$ block, \( \Ztb_j^{\mathsf{MHA}} \), is first normalized using $\mathsf{RMSNorm}$ and then processed by the $\mathsf{SwiGLU}$ block \eqref{eq:SwiGLU}. The transformed representation is combined with the residual $\mathsf{MHA}$ output \( \Ztb_j^{\mathsf{MHA}} \) \eqref{eq:MHAttn}, thereby enhancing representational capacity while preserving gradient stability.

\subsubsection{\textbf{Modifications Outside the Transformer Encoder Block}}
In the \emph{vanilla-T} model, a class token \( \mathbf{X}_{\text{cls}} \) is appended to the input sequence to capture the global representation, which is subsequently used for prediction.
However, the inclusion of the class token introduces an additional token to those generated during tokenization from PDP. Since the $\mathsf{MHA}$ layer exhibits quadratic complexity with respect to the number of tokens (\(\tilde{N}_{\rm tk}\)), this addition significantly increases the computational complexity. To address this issue, the class token mechanism is replaced with a Global Average Pooling (GAP) layer. The GAP layer, placed after the $N_{\rm L}$ Transformer encoder, replaces the class token mechanism by aggregating information from all tokens through averaging across the token dimension. This approach enables the model to utilize the collective representation for prediction while significantly reducing computational overhead. In this case, the total number of tokens in the embedding \(\mathbf{X}_{\text{emb}} = [\mathbf{X}_{\text{inp}}] \in \mathbb{R}^{\tilde{N}_{\rm tk} \times D_{\text{emb}}}\) becomes \(\tilde{N}_{\rm tk} = N_{\rm tk}\). The RMS normalization is applied on top of the aggregated information from the GAP layer, ensuring a well-normalized representation. This normalized output is then passed to the $\mathsf{IH}$ for the final prediction:
\begin{equation}
\begin{aligned}
\Ytb=\mathsf{IH}\left(\mathsf{RMSNorm}\left(\text{GAP} (\Ztb_{N_L}^{\text{ENC}})\right)\right)~.
\end{aligned}
 \label{eq:predSwish}
\end{equation}

Furthermore, due to the proposed tokenization approach described in Section V, positional embeddings are no longer required. In this approach, the entire sequence of sensor data is treated as a single token, and since the sensor positions are fixed, the positional order is inherently stored within the token itself. As a result, the positional embedding layer used in the \emph{vanilla-T} can be omitted, simplifying the model architecture and reducing computational overhead:
\begin{equation}
 \Ztb=\mathbf{X}_{\text{emb}}~.
\label{eq:PosEMod}
\end{equation}

An additional advantage of eliminating positional embeddings is that it prevents the introduction of bias into token representations, which can disrupt the modeling of multivariate correlations. 

The L-SwiGLU-T architecture integrates RMSNorm, SwiGLU feed-forward layers, GAP, and the removal of positional embeddings to form a domain-optimized Transformer variant for PDP-based indoor localization. Each modification is physically motivated: RMSNorm stabilizes wide dynamic-range inputs, SwiGLU selectively filters relevant multipath components, GAP reduces quadratic attention cost, and eliminating positional embeddings avoids introducing bias while simplifying the model structure.

\section{Evaluation and Analysis}\label{sec:Nume}
For wireless sensor applications, computation and energy resources are typically highly constrained. In this study, to evaluate the performance of the proposed model and tokenization approach, we consider small, medium, and large Transformer models, each with strict limits on the number of FLOPs \(mathcal{B}\): 4.5M, 16.5M, and 63.5M FLOPs, respectively. These constraints enable a comprehensive analysis of how data requirements and computational demands vary across models of different sizes. 

The remaining part of this section is structured as follows. First, we outline the training approach. Next, we provide a comparison of the proposed tokenization methods with the vanilla architecture. Finally, we evaluate the proposed \emph{L-SwiGLU-T} architecture, utilizing the \emph{SST} method, against the \emph{vanilla-T}.
% against both the vanilla transformer and the state-of-the-art ResNet model.
%When examining the specifications of the transformer model for the TRP Snapshot Tokenization method, as detailed in , it is important to note that the transformer models used for the other two tokenization approaches exhibit the same complexity as the one employed for the TRP Snapshot Tokenization approach.

\vspace{-3mm}
\subsection{Training Approach}
The availability of large datasets has been a major contributor to the success of DL models. However, collecting large quantity of RF data and the corresponding ground truth device locations is very costly and time-consuming. There is hence a strong need for effective data augmentation techniques to alleviate the data collection burden. We adopt the three RF data augmentation techniques introduced in \cite{khamesi2024} and summarize them briefly in the following.

\subsubsection{\textbf{Random Signal Dropping}} Out of the 18 received PDPs, we determine the number of PDPs to drop (set to all zeros) via
    \begin{equation}
    \begin{aligned}
            D=D_{\text{max}}\beta ~,
    \end{aligned}
     \end{equation}
     where $D_{\text{max}}$ is the maximum number of PDPs that can be dropped, which is set to 7, and $\beta$ is beta distributed with parameter set to (0.1,0.1). The beta distribution parameter is chosen such that no PDP is dropped roughly half the time and $D_{\text{max}}$ PDPs are dropped roughly half of the time. This emulates scenarios where some sensor nodes might be noisy, corrupted, or missing in real-world wireless applications.
     
\subsubsection{\textbf{Random Signal Shifting}} \label{random} Each of the 18 TD PDPs is independently shifted randomly along the time axis.
\begin{equation}
    \begin{aligned}
        \left\{ \overset{\leftrightarrow}{p} \left[ d \right] \right\}_d & \overset{\triangle}{=} \sum_{a=0}^{A-1} \bigg| \mathsf{IFFT}_{N_{\text{FFT}}} \left( \{ e^{-j 2 \pi k \Delta_f \delta} H_a[k]\}_{k=-\frac{N}{2}}^{\frac{N}{2}-1}  \right)\bigg|^2 ~,
    \end{aligned}
\end{equation}
where, \(\overset{\leftrightarrow}{p}[d]\) represents the delay-shifted PDP and $\delta$ follows a zero-mean truncated normal distribution with parameter $\sigma_{\text{RSS}}^2$ and limits of $(-2\sigma_{\text{RSS}},2\sigma_{\text{RSS}})$. This mimics the effect of network synchronization or transmit/receive timing offsets between the sensors and device, which are common impairments in real-world scenarios. We follow the suggested setting of $\sigma_{\text{RSS}}=25$ ns.
% \ref{random}
\subsubsection{\textbf{Smoothed Regression Mixup (SRM)}} This is an improved mixup approach for regression problems. Two data samples, A and B, with the dimension $18\times256$ PDPs, $x_A$ and $x_B$, are selected from a mini-batch to synthesize a new mixup PDP: $x_C=\lambda x_A+(1-\lambda)x_B$, based on the probability given by a Gaussian kernel $\exp \left( -\frac{d(A, B)}{2 \sigma_{\text{mix}}^2} \right)
$ \cite{yao2022c}. Here, $\lambda$ is a beta distributed random mixing ratio with parameter (2,2), $d(A, B)$ is a similarity measure between the two data samples, and $\sigma_{\text{mix}}^2$ is a hyperparameter. The probability kernel is designed such that two samples more similar to each other are more likely used as mixup inputs, while two rather different samples are rarely paired for mixing. Given the mixup PDP input of $x_C$, the loss function for the model output $\hat{y}_C$ is computed as the weighted average of the $L_1$ losses against the pre-mixing labels:
\begin{equation}
    \text{loss}_{\text{SRM}} = \lambda \| \hat{y}_C - y_A \|_1 + (1 - \lambda) \| \hat{y}_C - y_B \|_1~.
\end{equation}
To reduce complexity, we compute the similarly measure by $d(A, B) \overset{\triangle}{=} \| p_A - p_B \|_2^2$, where $p_A$ and $p_B$ are the device 2D coordinates, and set $\sigma_{\text{mix}}^2=4$. This results in SRM selecting $A=B$ with roughly $50\%$ probability. That is, roughly half of the data samples in a mini-batch are replaced with synthetic ones while the other half are unmodified by SRM.

% \begin{figure*}[t] % Use figure* to span both columns
%     \centering
%     \subfloat[$y=x$]{%
%         \includegraphics[width=0.3\textwidth]{Small_Model_allD.eps}
%         \label{fig:y_equals_x}}
%     \hfill
%     \subfloat[$y=3\sin x$]{%
%         \includegraphics[width=0.3\textwidth]{Medium_dataset_ALLM.eps}
%         \label{fig:three_sin_x}}
%     \hfill
%     \subfloat[$y=5/x$]{%
%         \includegraphics[width=0.3\textwidth]{Large_dataset_ALLM.eps}
%         \label{fig:five_over_x}}
%     \caption{Three simple graphs}
%     \label{fig:three_graphs}
%     \vspace{-5mm} % Adjust this value if you need to bring it up slightly
% \end{figure*}
% CDF of 2D positioning error on medium dataset.
\begin{figure*}[t] % Use figure* to span both columns
    \centering
    \subfloat[Small dataset.]{%
        \includegraphics[width=0.3\textwidth]{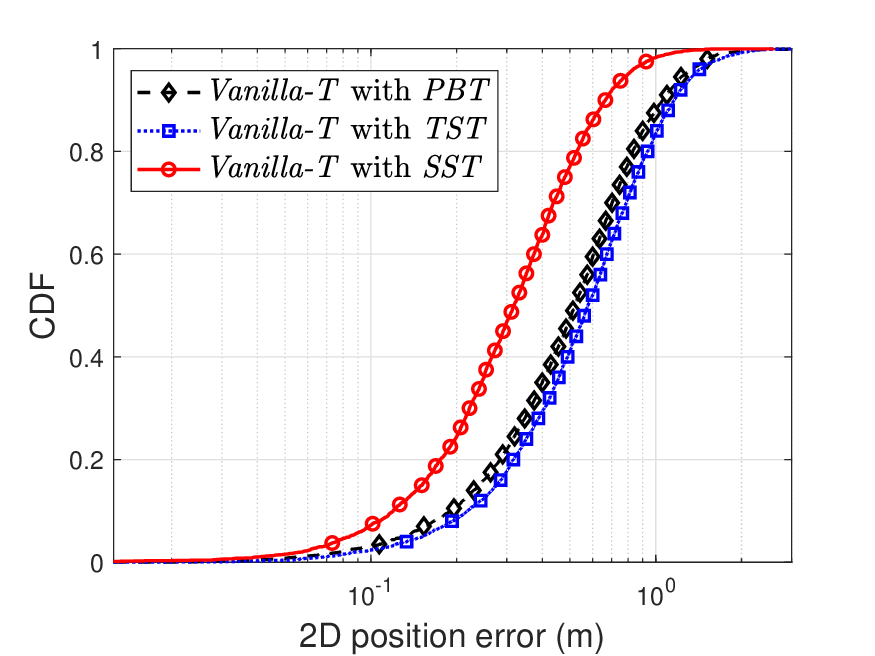}
        \label{fig:CA1}}
    \quad % or \hspace{0.05\textwidth}
    \subfloat[Medium dataset.]{%
        \includegraphics[width=0.3\textwidth]{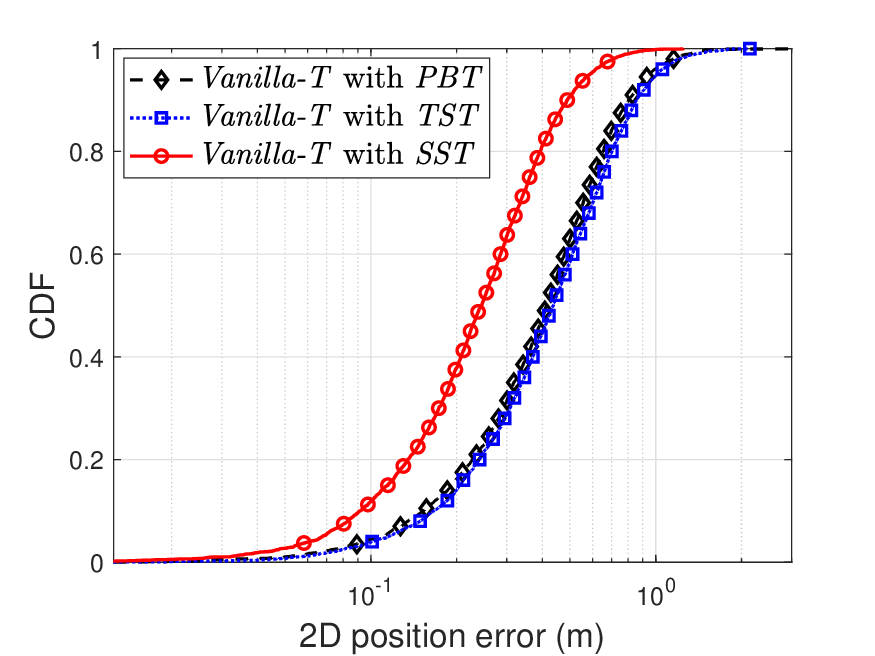}
        \label{fig:CA2}}
    \quad % or \hspace{0.05\textwidth}
    \subfloat[Large dataset.]{%
        \includegraphics[width=0.3\textwidth]{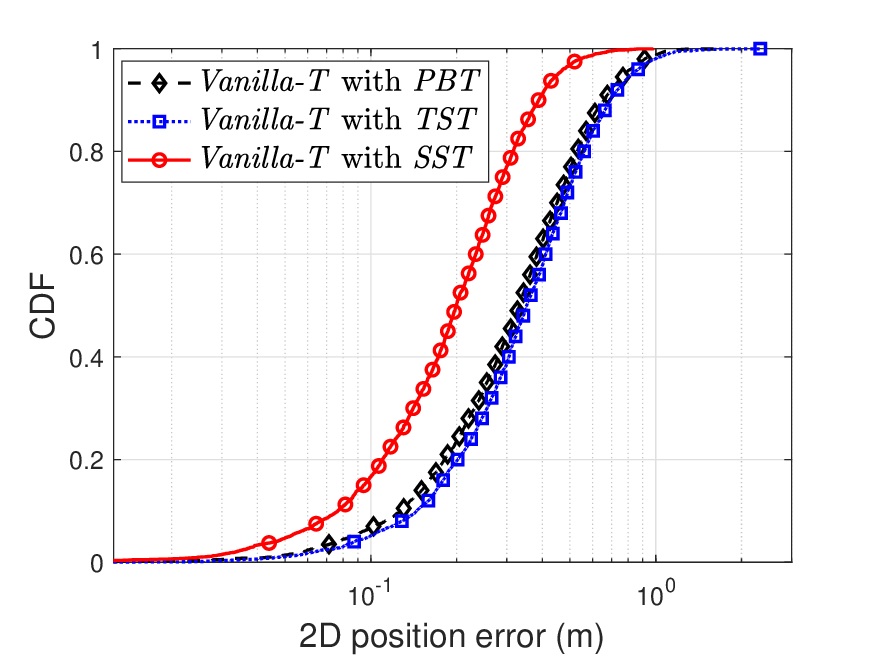}
        \label{fig:CA3}}
    \caption{Comparison of the CDF of 2D positioning errors for large models on different dataset sizes, evaluated across three tokenization methods using the \emph{Vanilla-T.}}
    \label{fig:CA}
    \vspace{-5mm} % Adjust this value if needed
\end{figure*}

\begin{figure*}[t] % Use figure* to span both columns
    \centering
    \subfloat[Small dataset.]{%
        \includegraphics[width=0.3\textwidth]{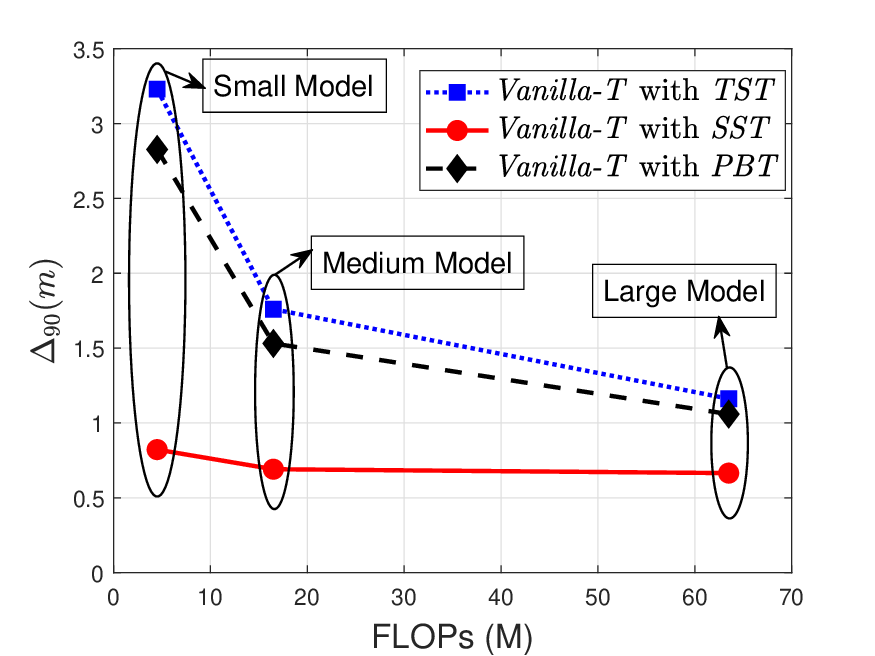}
        \label{fig:A1}}
    \quad % or \hspace{0.05\textwidth}
    \subfloat[Medium dataset.]
    {%
        \includegraphics[width=0.3\textwidth]{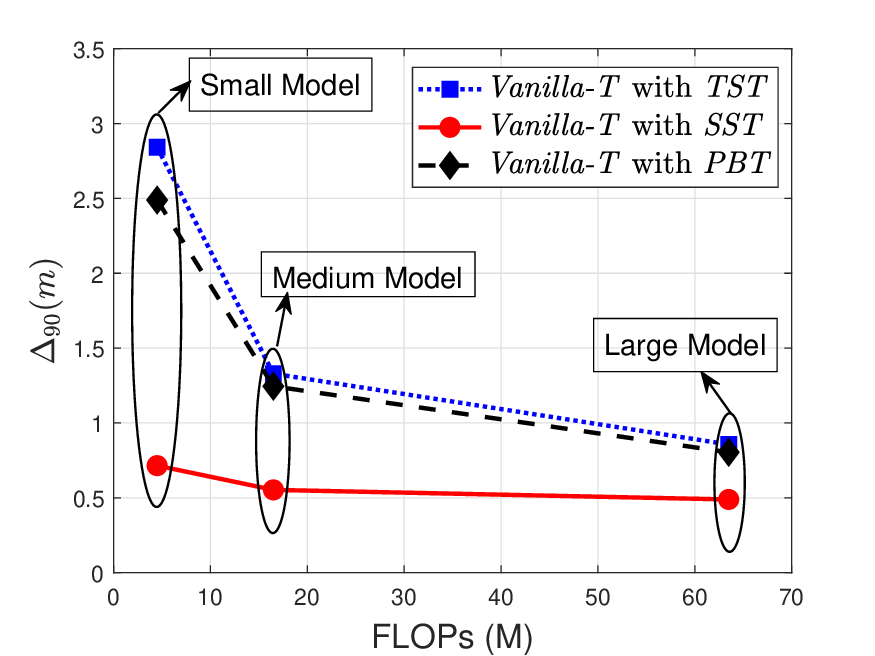}
        \label{fig:A2}}
    \quad % or \hspace{0.05\textwidth}
    \subfloat[Large dataset.]{%
        \includegraphics[width=0.3\textwidth]{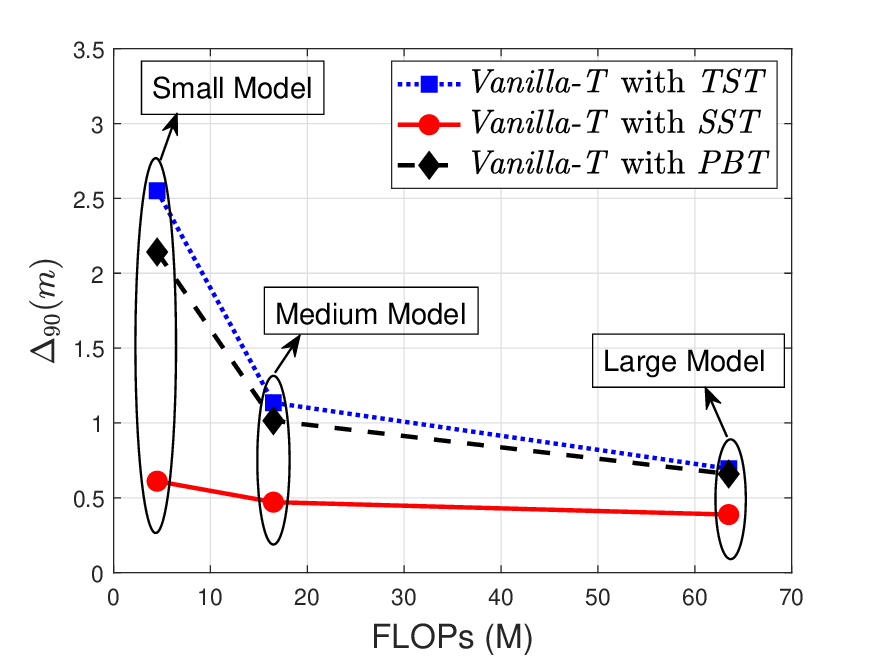}
        \label{fig:A3}}
    \caption{Comparison of \(\Delta_{90}\) positioning errors for varying model and dataset sizes, evaluated across three tokenization methods using the \emph{Vanilla-T}.}
    \label{fig:A}
    \vspace{-5mm} % Adjust this value if needed
\end{figure*}

For training, we use the JAX framework from Google and a batch size of $B=400$ PDPs sampled from the trainset with $40,000$ PDPs. Network parameters are updated by the Adam algorithm with its default settings and a weight decay factor of $0.001$. We use a cosine learning rate schedule with a warm-up that varies between $1e^{-5}$ and $2e^{-3}$ over $2000$ training epochs. All test results are evaluated using a separate set of $4,000$ test PDPs. Instead of selecting the model based on the best validation performance or using the weights from the last epoch, we use the Exponential Moving Average (EMA) of the model parameters with a decay rate of $\alpha=0.9$. This smooths out fluctuations in the parameters, leading to more stable and reliable weights. At each step, the EMA is updated as:

\begin{equation}
    \begin{aligned}
    \theta^{\text{EMA}}_{t} = \alpha \cdot \theta^{\text{EMA}}_{t-1} + (1 - \alpha) \cdot \theta_{t} ~,
    \end{aligned}
\end{equation}
where, \(\theta^{\text{EMA}}_{t}\) and \(\theta^{\text{EMA}}_{t-1}\) represent the EMA at epoch \(t\) and \(t-1\), while \(\theta_t\) denotes the model weight at epoch \(t\).
\subsection{Performance Analysis of the Proposed Tokenization Methods} \label{sub:vanilla}
The FLOPs complexity and model specifications are greatly influenced by the tokenization approach used. Table \ref{model} details the Transformer model specifications, with three values in each category being: \{$N_{\rm L}$, $D_{\text{emb}}$, $h_{\text{dim}}$\}. For instance, \{5,12,18\}, vanilla-T architecture consists of 5 encoder blocks with embedding dimension $D_{\text{emb}}=12$, and hidden dimension $h_{\text{dim}}=18$. The number of attention heads is fixed at $N_{\rm h}=6$, and \(\sigma\) is defined as the Rectified Linear Unit (ReLU) for all approaches. For the PBT framework, \( W_{\rm h} \) and \( W_{\rm w} \) are defined as 8 and 3, respectively, resulting in a total of $N_{\rm tk}=96$ tokens. Each token is represented as a vector of length $N_{\rm st}=24$.

% \begin{center}
% \captionof{table}{Transformer model configurations.}\label{modelspecs}
% \label{fig:tabel1} \label{model}
%  \begin{tabular}{||c c c c||}
%  \hline
%  Model & $N_{\rm L}$ & $D_{\text{emb}}$ & $h_{\text{dim}}$  \\ [0.5ex]
%  \hline
%  Small & \{5,3,6\} & \{12,12,48\}  & \{18,18,68\} \\
%  \hline
%  Medium & \{8,5,10\} & \{24,24,72\} & \{44,44,122\} \\
%  \hline
% Large & \{16,13,16\} & \{36,30,96\} & \{86,86,316\}  \\
%  \hline
% \end{tabular}
% \end{center}

 % Model & $N_{\rm L}$ & $D_{\text{emb}}$ & $h_{\text{dim}}$  \\ [0.5ex]
 
% \begin{center}
%  \small
% \captionof{table}{Transformer model configurations.}\label{modelspecs}
% \label{fig:tabel1} \label{model}
%  \begin{tabular}{||c c c c c||}
%  \hline
%  Model & $\emph{PBT}$ & $\emph{TST}$ & $\emph{SST}$ & FLOPs\\ [0.8ex]
%  \hline
%  Small & \{5,12,18\} & \{3,12,18\}  & \{6,48,68\} & 4.5 M \\
%  \hline
%  Medium & \{8,24,44\} & \{5,24,44\} & \{10,72,122\} & 16.5M \\
%  \hline
% Large & \{16,36,86\} & \{13,30,86\} & \{16,96,316\} & 63.5  \\
%  \hline
% \end{tabular}
% \end{center}

\begin{center} 
 \small
\renewcommand{\arraystretch}{1.3} % Adjust row height factor
\captionof{table}{Transformer model configurations.}\label{modelspecs} \label{model}
\begin{tabular}{||c c c c c||}
 \hline
 Model & $\emph{PBT}$ & $\emph{TST}$ & $\emph{SST}$ & FLOPs \\ [0.5ex]
 \hline
 Small & \{5,12,18\} & \{3,12,18\}  & \{6,48,68\} & 4.5 M \\
 \hline
 Medium & \{8,24,44\} & \{5,24,44\} & \{10,72,122\} & 16.5 M \\
 \hline
 Large & \{16,36,86\} & \{13,30,86\} & \{16,96,316\} & 63.5 M\\
 \hline
\end{tabular}

\end{center}

% \begin{center}
% \captionof{table}{Transformer model configurations.}\label{modelspecs}
% \begin{tabular}{||c p{1.2cm} p{1.2cm} p{1.4cm} c||} % Adjust width as needed
%  \hline
%  Model & $\emph{PBT}$ & $\emph{TST}$ & $\emph{SST}$ & FLOPs\\ [0.5ex]
%  \hline
%  Small & \{5,12,18\} & \{3,12,18\}  & \{6,48,68\} & 4.5 M \\
%  \hline
%  Medium & \{8,24,44\} & \{5,24,44\} & \{10,72,122\} & 16.5M \\
%  \hline
%  Large & \{16,36,86\} & \{13,30,86\} & \{16,96,316\} & 63.5 \\
%  \hline
% \end{tabular}
% \end{center}

% \textbf{Efficiency of TRP Snapshot Tokenization}
The performances of the three tokenization methods are evaluated by analyzing the Cumulative Distribution Function (CDF) of the 2D positioning error for the device, as illustrated in Fig.~\ref{fig:CA}. The figure presents the CDF results for a large \emph{Vanilla-T} model evaluated across small, medium, and large datasets (from left to right) using all three tokenization methods. It is observed that the proposed \emph{SST} tokenization method consistently outperforms the other two methods, namely \emph{PBT} and \emph{TST}, achieving a significant performance improvement in all three cases.
In Fig.~\ref{fig:CA3} \emph{SST} method outperforms other tokenization methods, achieving a 90th percentile ($\Delta_{90}$) error of $0.388$~m. In comparison, \emph{PBT} and \emph{TST} report $\Delta_{90}$ errors of $0.659$~m and $0.694$~m, respectively. These findings indicate that \emph{SST} effectively captures environmental dependencies and multivariate correlations, enhancing model performance, while the other methods fuse information from sensors, which leads to poor performance.

It can be observed that as the dataset size increases (from Fig.~\ref{fig:CA1} to Fig.~\ref{fig:CA3}), the CDF curve shifts to the left, indicating a reduction in positioning error. Specifically, for the \(\Delta_{90}\) error, the \emph{TST} tokenization method demonstrates a percentage improvement of 26.34\% when the dataset increases from small to medium, and $40.28$\% when increasing from small to large. Similarly, for the \emph{PBT} tokenization method, the percentage improvement is $24.08$\% for a small to medium dataset increase, and $37.77$\% for a small to large dataset increase. In comparison, the proposed \emph{SST} tokenization method exhibits a percentage improvement of $26.47$\% as the dataset grows from small to medium, and $40.75$\% when transitioning from small to large. These results highlight that while all tokenization methods benefit from larger datasets, the proposed \emph{SST} achieves superior improvements. Furthermore, the comparatively lower percentage improvements of \emph{TST} and \emph{PBT} suggest that their suboptimal tokenization strategies limit the model's ability to effectively leverage larger datasets, whereas \emph{SST} demonstrates its capability to harness the additional data for enhanced performance.

% \begin{figure}
% %\vspace{-0.5cm}
% 	\includegraphics[width=0.99\linewidth]{LM_LD_CDF.eps}
% 	\centering
% %\vspace{-0.3cm}
% 	\caption{CDF of 2D positioning error for all three tokenization methods using a large transformer model trained on the large dataset.}
% 	\label{Fig1}
% \vspace{-0.35cm}
% \end{figure} 
% \\
% \begin{figure*}[t] % Use figure* to span both columns
%     \centering
%     \subfloat[$y=x$]{%
%         \includegraphics[width=0.30\textwidth]{Small_Model_allD.eps}
%         \label{fig:y_equals_x}}
%     \hfill
%     \subfloat[$y=3\sin x$]{%
%         \includegraphics[width=0.30\textwidth]{Medium_dataset_ALLM.eps}
%         \label{fig:three_sin_x}}
%     \hfill
%     \subfloat[$y=5/x$]{%
%         \includegraphics[width=0.30\textwidth]{Large_dataset_ALLM.eps}
%         \label{fig:five_over_x}}
%     \caption{Three simple graphs}
%     \label{fig:three_graphs}
% \end{figure*}
% \textbf{Performance Analysis of Transformer Models on Small Dataset:} 
Fig.~\ref{fig:A} illustrates the $\Delta_{90}$ errors for small, medium, and large models across datasets of varying sizes. As shown in Fig.~\ref{fig:A1}, the $\Delta_{90}$ error for all model sizes trained on the small dataset indicates that the proposed \emph{SST} consistently achieves the lowest positioning errors. A similar trend is observed in Fig.~\ref{fig:A2} and \ref{fig:A3}, where the \emph{Vanilla-T} models are trained on medium and large datasets. Comparing the $\Delta_{90}$ error of the small, medium, and large models trained on the large dataset, the proposed tokenization method achieves percentage improvements of $76.06$\%, $58.53$\%, and $44.08$\%, respectively, over the \emph{TST}. Despite having the same computational complexity in terms of FLOPs, the performance gap between the models is notable, emphasizing the effectiveness of the proposed tokenization method.

\begin{figure}[t] % Add more flexible placement options
    \centering
    \includegraphics[width=0.85\linewidth]{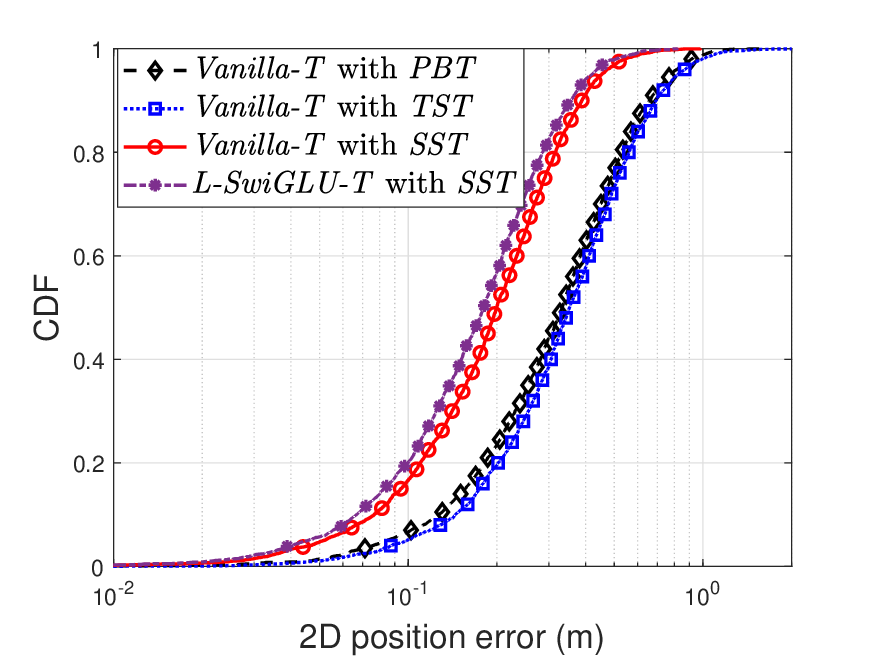}
    \caption{The CDF of 2D positioning errors for large \emph{L-SwiGLU} model trained on large dataset using \emph{SST} is shown alongside the CDFs of all tokenization methods already presented in Fig.~\ref{fig:CA3}.}
    \label{FigSGLUC}
    \vspace{-0.5cm} % Reduce or remove to minimize spacing
\end{figure}

% \begin{figure}
% %\vspace{-0.5cm}
% %\hspace{-0.8cm}
% \includegraphics[width=1\linewidth]{SGLU_LM_LD.eps}
% 	\centering
% %\vspace{-0.3cm}
% 	\caption{CDF of 2D positioning error for all three tokenization methods using a large transformer model trained on the large dataset.}
% 	\label{FigSGLUC}
% \vspace{-0.01cm}
% \end{figure} 

Considering model complexity, the model trained on a small dataset using \emph{SST} (see Fig.~\ref{fig:A1}) achieves superior positioning accuracy. Notably, the small model trained with the proposed \emph{SST} tokenization method not only outperforms models of similar size but also surpasses both medium and large models trained with \emph{PBT} and \emph{TST} tokenization techniques. Specifically, the small model using \emph{SST} achieves a $70.92$\% improvement over models with comparable computational complexity. Furthermore, it demonstrates a $46.31$\% improvement compared to a model $3.66$ times larger trained with \emph{PBT} and achieves a $22.38$\% improvement over a model $14.1$ times larger trained with \emph{PBT}. These results highlight that efficient tokenization can achieve higher accuracy with lower computational requirements, as also observed in Fig.~\ref{fig:A2} and \ref{fig:A3}.
% Percentage improvement for the small model across three datasets.

\begin{figure*}[b] % Use figure* to span both columns
    \centering
    \subfloat[Small model.]{%
        \includegraphics[width=0.3\textwidth]{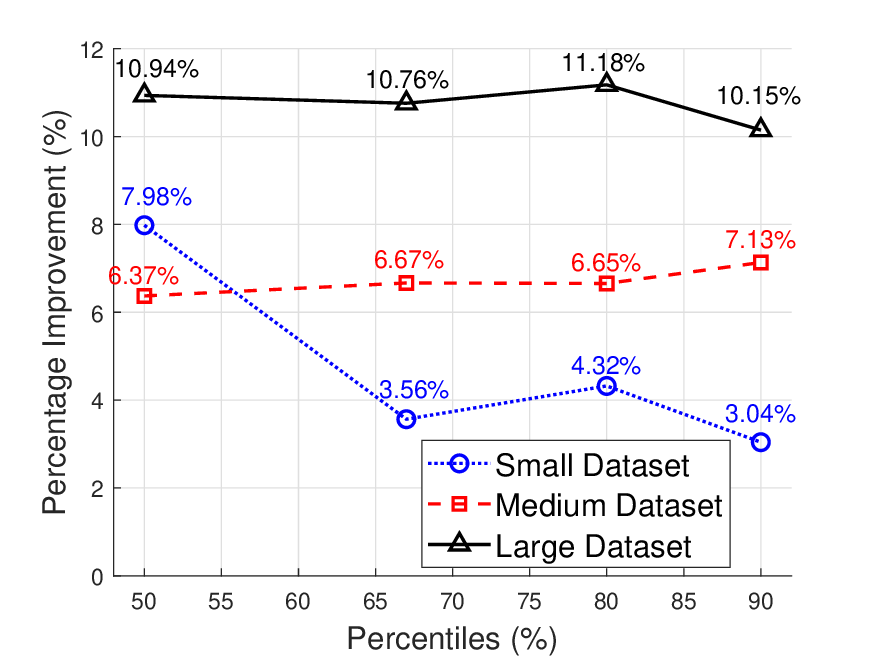}
        \label{fig:PA1}}
    \quad % or \hspace{0.05\textwidth}
    \subfloat[Medium model.]{%
        \includegraphics[width=0.3\textwidth]{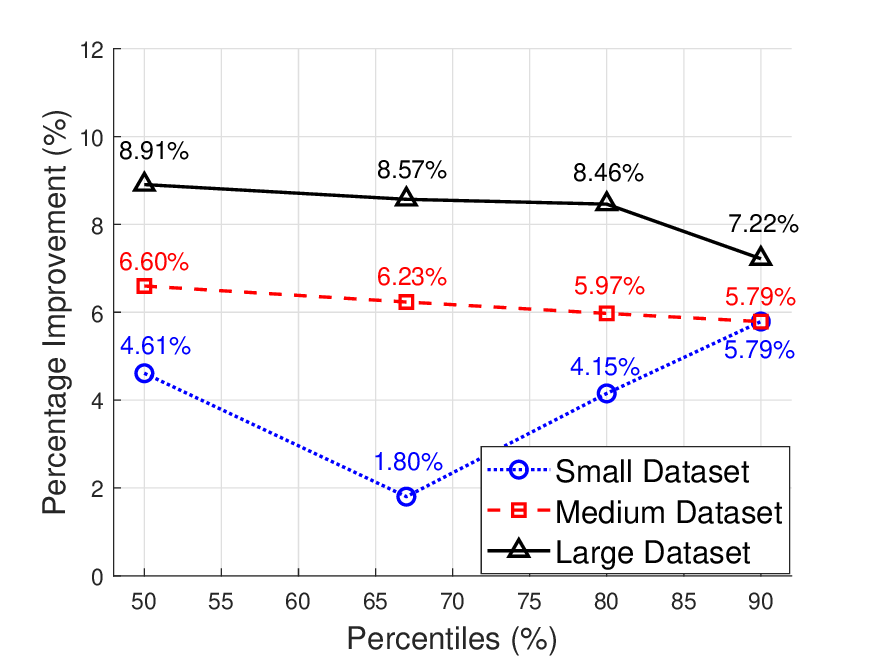}
        \label{fig:PA2}}
    \quad % or \hspace{0.05\textwidth}
    \subfloat[Large model.]{%
        \includegraphics[width=0.3\textwidth]{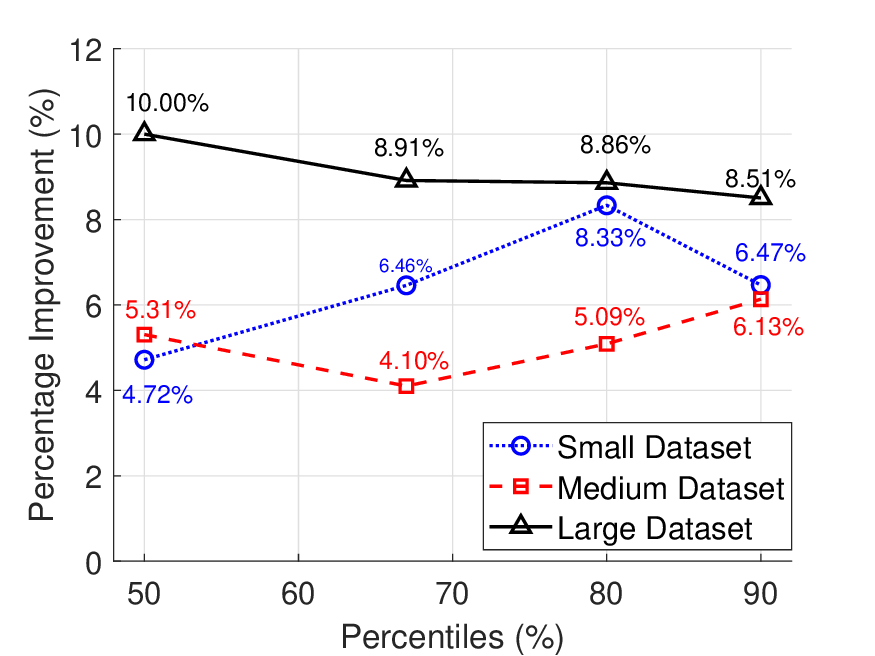}
        \label{fig:PA3}}
    \caption{Percentage improvement in 2D positioning errors of the proposed \emph{L-SwiGLU-T} over the \emph{Vanilla-T} using proposed \emph{SST} tokenization for small, medium, and large models across three datasets at the 50th, 67th, 80th, and 90th percentiles.}
    \label{fig:PA}
    \vspace{-5mm} % Adjust this value if needed
\end{figure*}

The $\Delta_{90}$ error for the large model trained on the small dataset using the proposed \emph{SST} tokenization method is $0.665$~m. In comparison, the $\Delta_{90}$ error for the large model trained on the large dataset is $0.694$~m with \emph{TST} and $0.659$~m with \emph{PBT}. Notably, the model trained on the small dataset using \emph{SST} outperforms the large dataset model using \emph{TST}, while achieving results comparable to those of the large dataset model using \emph{PBT}. These results demonstrate that the proposed \emph{SST} method significantly reduces the model's reliance on dataset size. 

In all three subfigures, it is evident that as the model size increases, the $\Delta_{90}$ error decreases. This trend is attributed to the larger models' enhanced capacity to capture the intricate variations and complex relationships within the dataset. Specifically, the $\Delta_{90}$ error of the proposed \emph{SST} approach for the small model trained on the small dataset is $0.882$~m, whereas the error for the small model trained on the large dataset reduces to $0.665$~m.  In contrast, when the large model is trained on the large dataset, the $\Delta_{90}$ error decreases to $0.388$~m. This suggests that the small model does not fully benefit from the larger dataset, likely due to the increased variability in the data, while the large model is better equipped to capture the complex behavior of the data more efficiently. Notably, for all model sizes, the proposed tokenization approach consistently outperforms others, demonstrating its effectiveness in reducing reliance on large datasets by efficiently capturing channel independence and multivariate correlations.

% \vspace{-0.3cm}
\subsection{Performance Analysis of the Proposed \emph{\emph{L-SwiGLU-T}}}

The proposed \emph{L-SwiGLU-T} model architecture shares a similar parameter configuration with the \emph{Vanilla-T} for \emph{SST} (Table. \ref{model}), differing primarily in the hidden dimension, denoted as \( h^{\text{SwiGLU}}_{\text{dim}} \). For the small, medium, and large models, the values of \( h^{\text{SwiGLU}}_{\text{dim}} \) are $54$, $94$, and $231$, respectively. These values are chosen such that the model complexity, in terms of FLOPs, remains equivalent to that of the \emph{Vanilla-T}, ensuring a fair comparison. The number of encoder blocks ($N_{\rm L}$), embedding dimension ($D_{\text{emb}}$), and attention heads ($H=6$) remain identical to those of the \emph{Vanilla-T}, as summarized in Table \ref{modelspecs}. As the proposed \emph{SST} tokenization method has consistently demonstrated superior performance across all cases, this subsection focuses exclusively on evaluating \emph{L-SwiGLU-T} using the \emph{SST} tokenization.

\begin{table}[t]
\centering
\caption{Performance summary: 2D localization error statistics in meters.}
\label{tab:performance_summary}
\begin{tabular}{|l|c|c|c|}
\hline
\textbf{Model} & \textbf{Mean} & \textbf{Std Dev} & \textbf{\(\Delta_{90}\)} \\
\hline
Vanilla-T + PBT & 0.3672 & 0.2152 & 0.6598 \\
Vanilla-T+ TST & 0.3953 & 0.2309 & 0.6940 \\
Vanilla-T + SST & 0.2203 & 0.1275 & 0.3880 \\
\textbf{L-SwiGLU-T + SST} & \textbf{0.1988} & \textbf{0.1181} & \textbf{0.3554} \\
\hline
\end{tabular}
\end{table}

% \begin{figure}
%     \centering
%     \subcaptionbox{}{\includegraphics[width=0.24\textwidth]{Percentage_impv_LM_AllD.eps}} 
%     \subcaptionbox{}{\includegraphics[width=0.24\textwidth]{Percentage_impv_LM_AllD.eps}} 
%     \subcaptionbox{}{\includegraphics[width=0.24\textwidth]{Percentage_impv_LM_AllD.eps}}
%     \subcaptionbox{}{\includegraphics[width=0.24\textwidth]{Percentage_impv_LM_AllD.eps}}
%     \caption{(a) blah (b) blah (c) blah (d) blah}
%     \label{fig:foobar}
% \end{figure}

% \documentclass{article}
% \usepackage{graphicx}
% \usepackage{caption}
% \usepackage{subcaption} % Use this instead of subfig to avoid conflicts
Fig.~\ref{FigSGLUC} presents the CDF of 2D positioning errors for the proposed \emph{L-SwiGLU-T} with \emph{SST} tokenization and compares it against the \emph{Vanilla-T} utilizing all three tokenization methods: \emph{PBT, TST,} and \emph{SST}. The results correspond to the scenario where a large model is trained on a large dataset. Despite maintaining similar computational complexity, the proposed \emph{L-SwiGLU-T} achieves notably better performance, outperforming the \emph{Vanilla-T}. Specifically, the $\Delta_{90}$ positioning error (\(\Delta_{90}\)) for the \emph{Vanilla-T} is $0.388$~m, whereas the proposed \emph{L-SwiGLU-T} achieves a reduced error of $0.355$~m, demonstrating its superior accuracy and efficiency in large-scale scenarios.  

To further strengthen the statistical validity of our results, we report the mean, standard deviation, and 90th percentile of the 2D localization error for each method in Table~\ref{tab:performance_summary}. As seen, the proposed \emph{L-SwiGLU-T} with \emph{SST} achieves the lowest average error of $0.1988$~m with the smallest standard deviation (Std Dev) of 0.1181~m, indicating consistent and accurate localization. This supports the observed CDF trends and addresses concerns about marginal gains being within the noise range.

The Fig.~\ref{fig:PA} demonstrates the percentage improvement in 2D positioning errors achieved by the \emph{L-SwiGLU-T} over the \emph{Vanilla-T} with the proposed \emph{SST} method across small, medium, and large models for three dataset sizes (small, medium, and large) at the 50th, 67th, 80th, and 90th percentiles. It can be seen that the proposed \emph{L-SwiGLU-T} model demonstrates a positive percentage improvement, indicating its superior performance over the \emph{Vanilla-T}. Across all model configurations (Fig.~\ref{fig:PA1} to Fig.~\ref{fig:PA3}), the large dataset consistently delivers the most significant performance gains, as it provides richer and more diverse training samples, enabling the model to better learn complex patterns and generalize effectively.

For the large dataset, the small model demonstrates a percentage improvement ranging from $10.15$\% to $11.18$\%, while the medium model shows a decrease to $8.91$\% to $7.22$\%. For the large model, the improvement ranges from $10.08$\% to $8.51$\%. The higher percentage improvement observed in the small model is attributed to its previously limited ability to fully utilize the large dataset when trained with the \emph{Vanilla-T}. With the proposed \emph{L-SwiGLU-T} model, the small model is now able to effectively leverage the large dataset, resulting in significant performance gains.

To assess the contribution of each architectural component in L-SwiGLU-T, we performed ablation experiments on the small model trained on the large dataset (Fig.~\ref{fig:PA1}). Replacing LayerNorm with RMSNorm improved training stability and efficiency. Removing the class token reduced complexity and yielded a 3.4\% accuracy gain. Eliminating positional embeddings avoided bias and improved accuracy by 4\%, while introducing the SwiGLU feedforward block contributed an additional 2.6\% improvement. These results highlight the effectiveness of each modification.

% For the small model, the improvement ranges from 10.15\% to 11.18\%, while for the medium model, it ranges from 8.91\% to 7.22\%. Similarly, for the large model, the improvement ranges from 10.08\% to 8.51\%. These results highlight that the proposed L-SwiGLU Transformer benefits significantly from larger datasets, though it is also outperforming in the case of small and medium datasets.
For the small and medium models (Fig.~\ref{fig:PA1}, and Fig.~\ref{fig:PA2}), improvements are moderate for the medium dataset and highest for the large dataset, while the small dataset yields limited gains. For the large model, the performance improvement on the small dataset is minimal at the 50th and 67th percentiles. However, at the (\(\Delta_{90}\)), the proposed \emph{L-SwiGLU-T} demonstrates superior performance. As model size increases, improvement becomes more sensitive to dataset size, reinforcing the need for appropriately scaled datasets to realize the full potential of larger models.

\begin{figure}
%\vspace{-0.5cm}
	\includegraphics[width=0.99\linewidth]{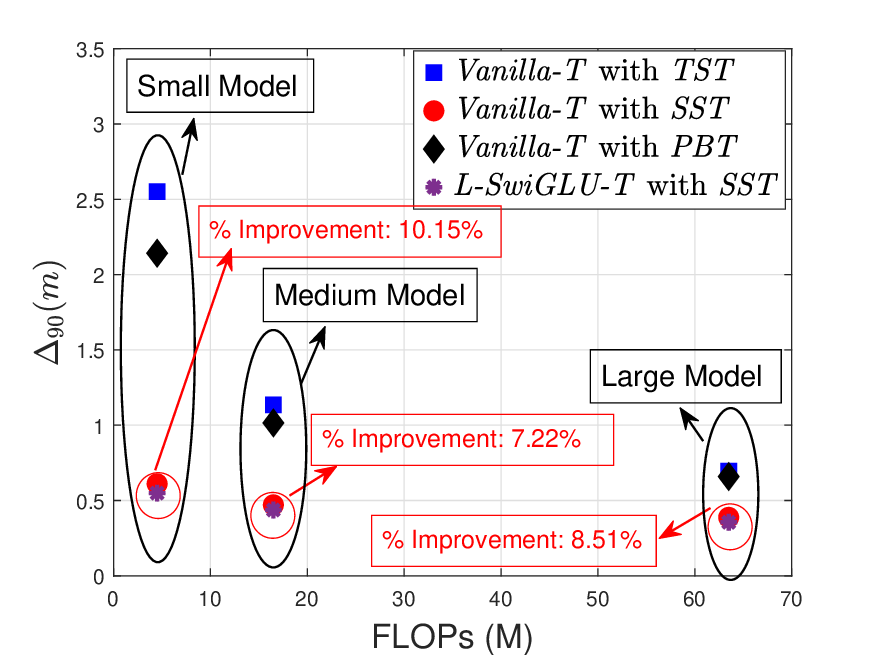}
	\centering
%\vspace{-0.3cm}
	\caption{Comparison of \(\Delta_{90}\) positioning errors for the \emph{L-SwiGLU-T} across varying model sizes on the large dataset, with the \emph{Vanilla-T} using three tokenization methods. }
	\label{FigSGLU90}
\vspace{-0.35cm}
\end{figure}  
Fig.~\ref{FigSGLU90} presents the \(\Delta_{90}\) positioning error for the proposed \emph{L-SwiGLU-T} using \emph{SST} compared to the \emph{Vanilla-T} across different tokenization methods on the large dataset. The \emph{L-SwiGLU-T} with \emph{SST} consistently achieves superior performance. Despite having similar computational complexity, the percentage improvement of the \emph{L-SwiGLU-T} over the \emph{Vanilla-T} (both using \emph{SST}) ranges from $7.22$\% to $10.15$\%. Compared to the \emph{Vanilla-T} using \emph{TST}, the percentage improvements are $78.48$\%, $61.50$\%, and $48.85$\% for the small, medium, and large datasets, respectively. Similarly, improvements over the \emph{Vanilla-T} with \emph{PBT} are $74.37$\%, $56.95$\%, and $46.13$\%. 
The proposed \emph{L-SwiGLU-T} demonstrates superior performance compared to the \emph{Vanilla-T} models with \emph{PBT} of significantly larger sizes. Specifically, the small \emph{L-SwiGLU-T} outperforms the medium \emph{Vanilla-T} trained with \emph{PBT}, which is $3.75$ times larger, achieving a $66.99$\% improvement. It also surpasses the large \emph{Vanilla-T} with \emph{PBT}, which is $14.1$ times larger, with a $46.13$\% improvement. When compared to the \emph{Vanilla-T} with \emph{SST}, the improvements over \emph{PBT} are $46.31\%$ and $22.38\%$, respectively, as detailed in Subsection \ref{sub:vanilla}. These results highlight the substantial performance improvements achieved by the proposed SwiGLU, with percent gains significantly increasing from $46.31\%$ to $66.99\%$ and from $22.38\%$ to $46.13\%$.
These results highlight the effectiveness of the proposed \emph{SST} tokenization and the significant performance gains achieved by the \emph{L-SwiGLU-T}.

\subsection{Attention Map Analysis of Proposed \emph{L-SwiGLU}-T Model}
To interpret how the proposed \emph{L-SwiGLU-T} model utilizes information from distributed sensors under NLOS conditions, we visualize the average attention weights across sensors from selected Transformer layers. Since each layer contains multiple attention heads that capture different representation aspects, we compute the mean attention scores across all six heads to obtain a more interpretable view.

\begin{figure*}[t] % Use figure* to span both columns
    \centering
    \subfloat[First Transformer layer]{%
        \includegraphics[width=0.31\textwidth]{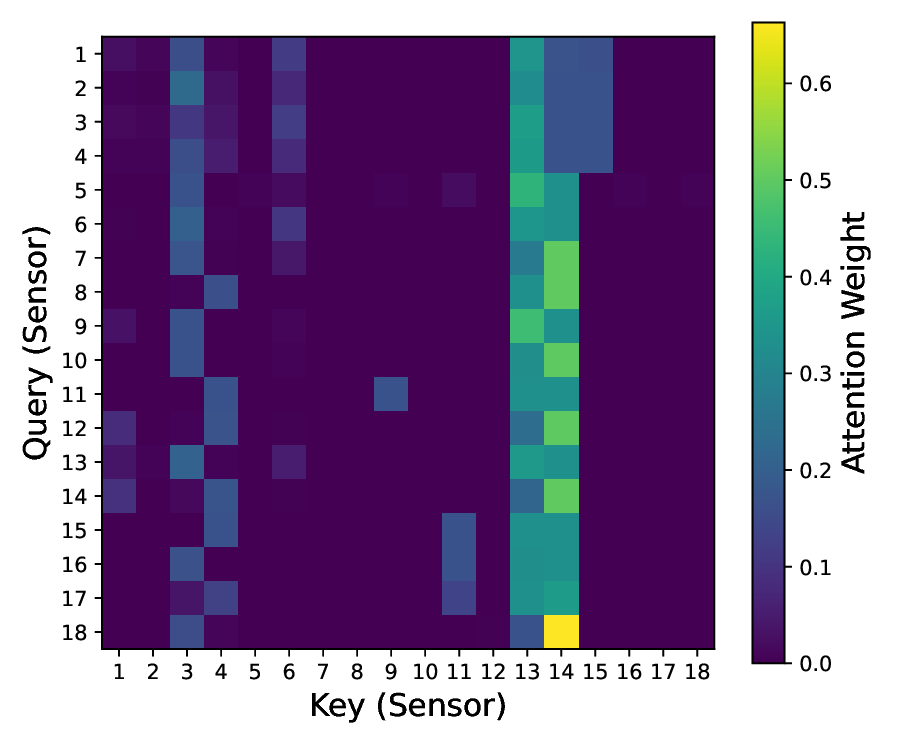}
        \label{fig:HM1}}    \quad % or \hspace{0.05\textwidth}
    \subfloat[Last Transformer layer]{%
        \includegraphics[width=0.31\textwidth]{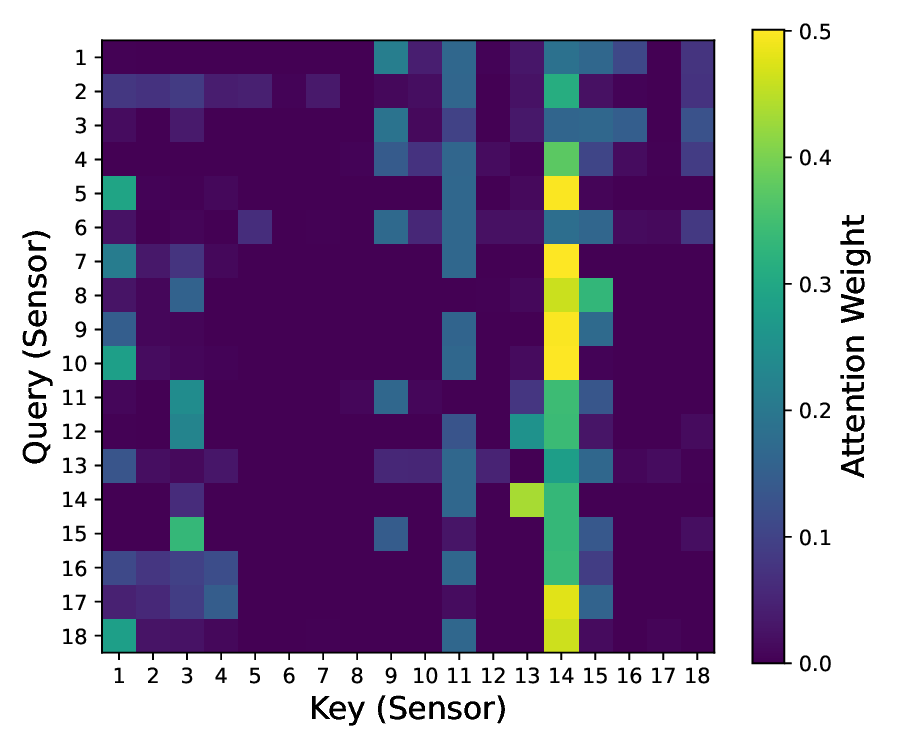}
        \label{fig:HM2}}
    \quad % or \hspace{0.05\textwidth}
    \subfloat[PDPs of selected sensors.]{%
        \includegraphics[width=0.31\textwidth]{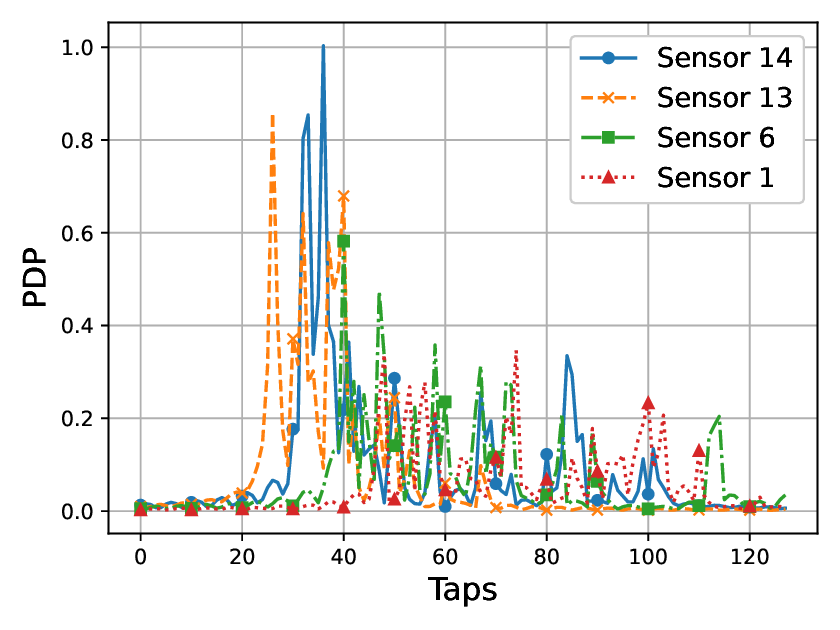}
        \label{fig:PDP1}}
    \caption{Visualization of attention and signal characteristics under an NLOS scenario. (a)–(b) show average attention weights across sensors from the first and last Transformer layers, respectively, while (c) presents the corresponding PDPs for selected sensors.}
    \label{fig:Attention}
    \vspace{-5mm} % Adjust this value if needed
\end{figure*}

Fig.~\ref{fig:HM1} and Fig.~\ref{fig:HM2} show the average attention weights from the first and last transformer layers for the small model on the small dataset, respectively, aggregated over six heads. It is evident that both early and deep layers assign significantly higher attention to Sensor 13 and Sensor 14, while suppressing contributions from less informative sensors such as Sensor 1.  To further support this behavior, we examine the PDPs of some sensors in Fig.~\ref{fig:PA}. The absence of early, high-power taps and the dispersion of energy across mid-to-late taps across all sensors is a strong indicator of the NLOS nature of the environment. Sensors 13 and 14 exhibit strong peaks at higher taps/delay-bins (e.g., taps 30–45), indicating dominant reflected paths arriving after significant delay. Sensor 1 shows a relatively delayed and diffuse profile, with no dominant early peak, consistent with deep NLOS or heavy obstruction. The attention mechanism clearly learns to assign higher weight to sensors with higher signal quality and lower multipath delay spread, reinforcing the model's capacity to dynamically prioritize informative spatial observations. This validates the model’s interpretability and effectiveness in leveraging distributed sensor data under multipath-rich indoor environments.

\subsection{Comparison of L-SwiGLU-T with Lightweight Transformer Baselines} \label{Light}
To benchmark efficiency and performance, we compare \emph{L-SwiGLU}-T with recent state-of-the-art lightweight Transformer architectures. The One Wide Feedforward is All You Need \emph{(OWFF)} Transformer \emph{(OWFF-T)} model [1] reduces redundancy by removing separate FFNs per encoder layer and instead employs a single, wide FFN shared across all layers. In contrast, \emph{DELIGHT-T} (Deep and Light-weight Transformer) [2] introduces a highly compact architecture that employs Group Linear Transformations to capture local representations. To incorporate global context, it applies feature shuffling across groups, analogous to channel shuffling in CNNs.

\begin{figure}
%\vspace{-0.5cm}
	\includegraphics[width=0.9\linewidth]{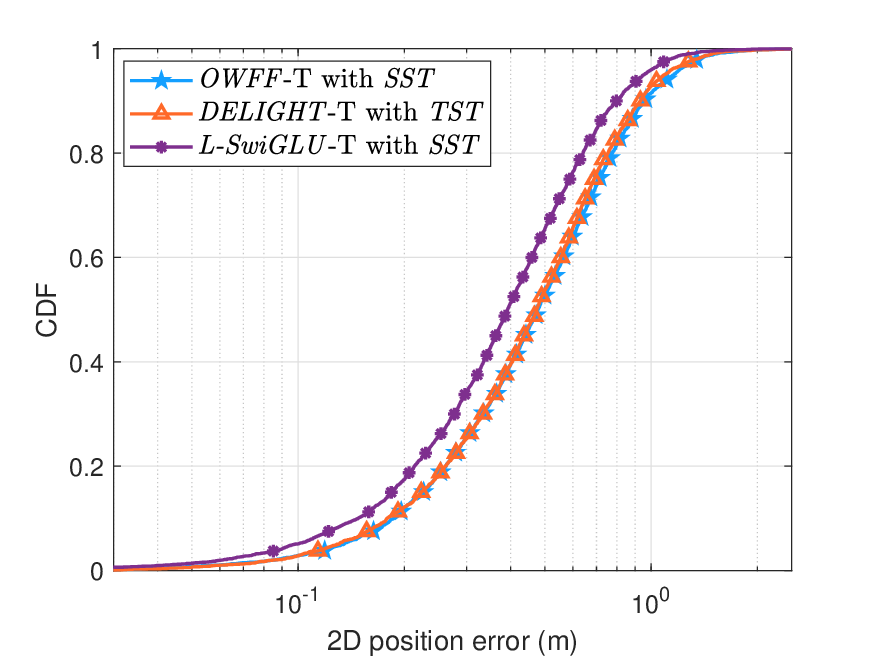}
	\centering
%\vspace{-0.3cm}
	\caption{CDF of 2D localization errors for \emph{L-SwiGLU-T}, \emph{OWFF}-T, and \emph{DELIGHT}-T using small model on the small dataset.}
	\label{LightR}
\vspace{-0.35cm}
\end{figure}

Fig.~\ref{LightR} plots the CDF of 2D localization errors for the proposed \emph{L-SwiGLU}-T, \emph{OWFF}-T, and \emph{DELIGHT}-T models on the small dataset. All models are constrained to 4.5M FLOPs for fair comparison. The proposed \emph{L-SwiGLU}-T demonstrates superior performance, achieving a \(\Delta_{90}\) error of \(0.7974\)~m, compared to \(0.9316\)~m for DELIGHT-T and \(0.9562\)~m for OWFF-T. This performance gain is attributed to its effective balance between computational efficiency and parameter utilization, leveraging gated activations and structural sparsity to enhance representational capacity without increasing model complexity.

\subsection{Performance Comparison of \emph{L-SwiGLU}-T Model Sizes on Real-World Dataset}
To demonstrate the practical effectiveness of the proposed \emph{L-SwiGLU-T} model, we evaluated it on the Ultra-Dense MaMIMO CSI Dataset \cite{nr6k-8r78-21}, which provides real-world channel measurements collected using a multi-antenna testbed. The dataset was recorded at a center frequency of 2.61~GHz with a 20~MHz bandwidth and $N = 100$ subcarriers. Among the available antenna configurations, we focused on the distributed antenna array (DA) topology, where $N_S = 8$ sensor nodes are placed at distinct locations surrounding the area of interest. Each sensor node is equipped with $A = 8$ receiving antennas, forming a spatially distributed sensing setup. The collected frequency-domain CSI was transformed into TD CIRs using \eqref{eq:TDCIR} with $N = 128$, followed by computation of the PDP via \eqref{eq:pdp}. Only the first $N_{\text{ts}} = 128$ delay taps were retained for evaluation.

For real-world evaluation using the \emph{SST} framework, the number of tokens is  $N_{\rm tk} = N_{\rm S} = 8$, with each token representing one of the eight distributed sensors, and each token having a feature dimension of $N_{\rm st} = N_{\textrm{ts}} = 128$. While our simulation studies considered $N_{\rm S} = 18$ sensors (see Section~\ref{SST}), the proposed \emph{L-SwiGLU-T} model is designed to support varying sensor counts. Only the number and embedding of input tokens need to be adapted to the deployment; the core model architecture remains unchanged. This demonstrates the flexibility of our SST-based formulation in handling diverse sensor configurations. Furthermore, spatial differences in sensor placement are implicitly learned through the attention mechanism, and the \emph{L-SwiGLU-T}’s ability to process variable-length sequences enables generalization to dynamic sensor layouts, even without retraining. Another observation is that in this dataset, the number of tokens is reduced to $N_{\rm tk} = N_{\rm S} = 8$, compared to 18 in the simulation setup. Since Transformer complexity scales quadratically with the number of tokens, this results in a substantial reduction in computational cost. Specifically, the FLOPs (\(\mathcal{B}\)) for the small, medium, and large models decrease from 4.5M to 1.8M, 16.5M to 7M, and 63.5M to 27.5M, respectively, highlighting the efficiency of the proposed model in real-world scenarios with fewer sensors.

\begin{figure}
%\vspace{-0.5cm}
	\includegraphics[width=0.9\linewidth]{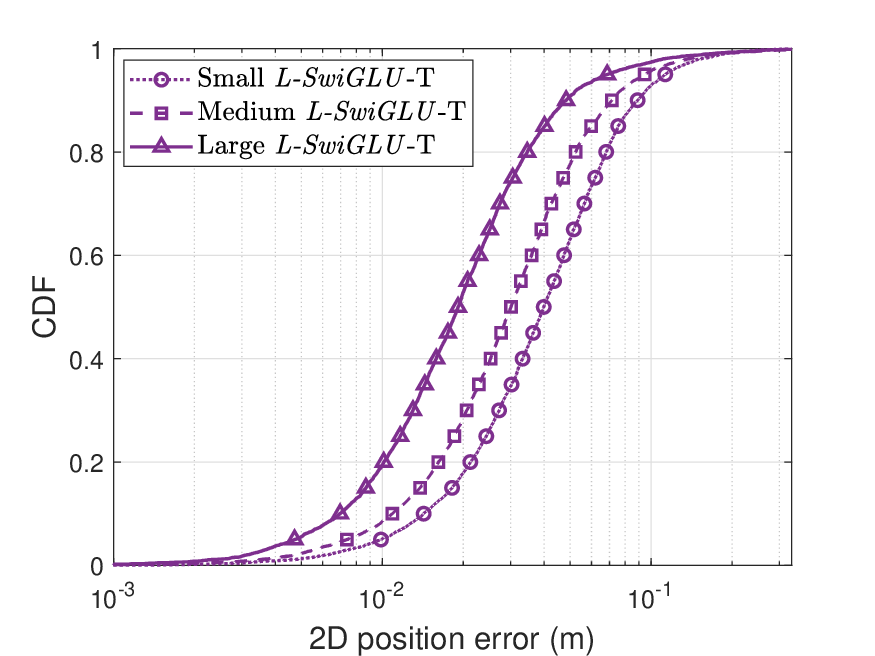}
	\centering
%\vspace{-0.3cm}
	\caption{CDF of 2D localization error for small, medium, and large \emph{L-SwiGLU-T} models on real world dataset.}
	\label{Realworld}
\vspace{-0.35cm}
\end{figure}

The real-world dataset contains approximately 250K samples; however, for this study, we utilize a subset of 40K samples, which we refer to as the large dataset in the context of this work. The CDF of the 2D localization error for the small, medium, and large versions of the proposed \emph{L-SwiGLU-T} is shown in Fig. \ref{Realworld}. As model capacity increases, the error consistently decreases, highlighting the model's ability to capture complex spatial features more effectively. Specifically, the 90th percentile errors are $0.0890$~m for the small model, $0.0714$~m for the medium model, and $0.0483$~m for the large model.

To benchmark against prior work, we compare our model's performance with the CNN approach implemented by the authors of the dataset in \cite{de2020csi}. Their CNN model achieved a mean absolute error (MAE) of $0.0823$~m, trained on the entire dataset of 250K samples. In contrast, our proposed \emph{L-SwiGLU-T} model was trained on a significantly smaller subset of only 40K samples, yet it consistently outperformed the CNN-based based across all model sizes. This highlights the efficiency and generalization capability of the Transformer-based architecture, even with limited data. The results are summarized in Table \ref{realworldt}.

\begin{table}[h]
\centering
\caption{MAE comparison: Proposed \emph{L-SwiGLU-T} vs. CNN baseline \cite{de2020csi}}
\label{realworldt}
\begin{tabular}{|c|c|}
\hline
\textbf{Model} & \textbf{MAE (m)} \\
\hline
CNN (Baseline, 250K samples) \cite{de2020csi} & 0.0823 \\
\emph{L-SwiGLU-T} Small (40K samples) & 0.0482 \\
\emph{L-SwiGLU-T} Medium (40K samples) & 0.0387 \\
\emph{L-SwiGLU-T} Large (40K samples) & 0.0267 \\
\hline
\end{tabular}
\end{table}

These results demonstrate that even under constrained data regimes, the proposed model effectively captures spatiotemporal features with higher precision than convolutional baselines, benefiting from attention mechanisms that model long-range dependencies across distributed sensors.

The proposed SST-based L-SwiGLU-T model is inherently flexible and not restricted to fixed input dimensions such as the number of sensors or PDP length. It readily accommodates different input sizes through token-level adaptation without requiring architectural modifications. As demonstrated in both simulation (with 18 sensors at 3.5 GHz/100 MHz) and real-world evaluation (with 8 sensors at 2.61 GHz/20 MHz), the same model generalizes effectively across diverse sensor configurations and frequency bands, underscoring its robustness for practical deployment. Multi-floor localization and cross-domain adaptation can be effectively achieved using transfer learning, where a pre-trained model can be fine-tuned with minimal data to adapt to new floors or environments.

\section{Conclusion} \label{Conclusion}
In this paper, we investigated the use of Transformer architectures for indoor localization in highly NLOS environments, emphasizing the often-overlooked role of tokenization in wireless signal modeling. We proposed a novel \emph{SST} method that effectively captures inter-sensor dependencies from PDP data while reducing computational overhead and data dependence. Complementing this, the lightweight \emph{L-SwiGLU-T} model integrates gated linear units, RMSNorm, and GAP to enhance feature selectivity and reduce computational cost. Together, these design elements significantly reduce FLOPs while maintaining high localization accuracy. Extensive evaluations demonstrated that the proposed framework achieves state-of-the-art accuracy, reducing the 90th-percentile localization error to $0.388$~m and outperforming models up to 14$\times$ larger with significantly fewer FLOPs and training samples. When tested on real-world datasets, the proposed model maintained strong generalization, achieving over 40\% improvement compared to CNN baselines. Overall, this work highlights the potential of domain-optimized Transformers for efficient and robust indoor localization. Future research will explore further architectural simplifications and hybrid tokenization strategies to enhance scalability and real-time deployment in practical wireless systems.

%Specifically, Transformer models trained on small datasets achieve superior performance compared to conventional approaches trained on significantly larger datasets.
\vspace{-0.2cm}

\bibliographystyle{IEEEtran}
\bibliography{mybib}

\end{document}